\documentclass[12pt]{article}

\usepackage{scicite}
\usepackage{amsmath}
\usepackage{amsfonts}
\usepackage{graphicx}
\usepackage{times}
\usepackage{xcolor}
\usepackage{soul}
\usepackage{color}
\usepackage{placeins}

\usepackage[margin=1.0in]{geometry}

\usepackage[font={small,stretch=1}, labelfont=bf]{caption} 

\title{Landslide Susceptibility Modeling by Interpretable Neural Network}

\author
{Khaled Youssef,$^{1\ast,5}$ Kevin Shao,$^{2\ast}$, Seulgi Moon,$^{2}$ Louis-S. Bouchard$^{1,3,4}$\\
\\
\normalsize{$^{1}$Department of Chemistry and Biochemistry, UCLA.}\\
\normalsize{$^{2}$Department of Earth, Planetary, and Space Sciences, UCLA.}\\
\normalsize{$^{3}$Department of Bioengineering, UCLA.}\\
\normalsize{$^{4}$California NanoSystems Institute, UCLA.}\\
\normalsize{$^{5}$Current address: Krannert Cardiovascular Research Center, University of Indiana School of Medicine.}\\
\\
\normalsize{$^{\ast}$ co-first authors} \\
 \\
\normalsize{Correspondence: S.G. Moon (sgmoon@ucla.edu), L.-S. Bouchard (lsbouchard@ucla.edu)} \\
\\
}

\makeatletter
\let\saved@includegraphics\includegraphics
\AtBeginDocument{\let\includegraphics\saved@includegraphics}
\renewenvironment*{figure}{\@float{figure}}{\end@float}
\makeatother

\date{}

\begin{document} 

\baselineskip24pt


\maketitle 

%

\begin{abstract}
 Landslides are notoriously difficult to predict because numerous spatially and temporally varying factors contribute to slope stability. Artificial neural networks (ANN) have been shown to improve prediction accuracy but are largely uninterpretable. Here we introduce an additive ANN optimization framework to assess landslide susceptibility, as well as dataset division and outcome interpretation techniques. We refer to our approach, which features full interpretability, high accuracy, high generalizability and low model complexity, as superposable neural network (SNN) optimization. We validate our approach by training models on landslide inventory from three different easternmost Himalaya regions.  Our SNN outperformed physically-based and statistical models and achieved similar performance to state-of-the-art deep neural networks. The SNN models found the product of slope and precipitation and hillslope aspect to be important primary contributors to high landslide susceptibility, which highlights the importance of strong slope-climate couplings, along with microclimates, on landslide occurrences.
\end{abstract}

\noindent {\bf\textsf{Introduction}}

Landslides are a major natural hazard that cause billions of dollars in direct damages and thousands of deaths globally each year~\cite{petley2012global,froude2018global}. 
Landslides can also cause various secondary hazards, such as damming and flooding, which often leave a region prone to subsequent damage following the initial event~\cite{huang2013landslide}.  Additionally, landslide debris may cause instability by perturbing river sedimentation and disrupting ecosystems~\cite{fan2019earthquake,huang2013landslide}.  As landslide hazards are expected to increase due to climate change, scientists have sought to more accurately assess landslide susceptibility~\cite{bui2012landslide,tien2019shallow,phong2019landslide,dikshit2020pathways,kirschbaum2020changes,stanley2017heuristic}, an estimate of the probability that a landslide may occur in a specific area, with the goal of mitigating the impact of landslides on the economy, public safety, and local ecosystems. 

Landslide occurrences are influenced by various factors including physical attributes of the terrain, such as slope, relief, and drainage areas, and material properties such as the density and strength of soil and bedrock~\cite{Dietrich1995,Montgomery1994, Montgomery1998, radbruch1982landslide}.  Also, environmental conditions such as climate, hydrology, ecology, and ground motion due to earthquakes may contribute to slope instability~\cite{guzzetti1999,Baum2002,Menuier2008}.   Landslide susceptibility is calculated from these various controlling factors either through physically-based models~\cite{Montgomery1994,Montgomery1998,Baum2002,Baum2010}, data-driven approaches utilizing statistical analysis~\cite{Lee2006,Regmi2014}, or machine learning techniques (ML), including random forest, support vector machines, and deep neural networks (DNN)~\cite{tien2019shallow,van2020spatially,bui2020comparing,lsnn1,lsnn2,lee2004determination, STANLEY2020104692}.

While substantial work has been devoted to assessing susceptibility, each model has shortcomings. Physically- or mechanistically-based approaches, based on the equilibrium between driving and resisting forces, have been widely applied to assess slope stability~\cite{Montgomery1994,Dietrich1995,Montgomery1998,Dietrich2001}. However, mechanistic models have limitations, including a limited number of variables, simplified assumptions of landslide geometry and certain environmental conditions (e.g., antecedent moisture, bedrock structure), and the high cost of geotechnical exploration necessary to estimate and calibrate for accurate subsurface properties (e.g., cohesive strength, pore pressure, weathering profile)~\cite{guzzetti1999}. Alternatively, data-driven approaches, including statistical and ML methods, can handle a large number of controls to assess susceptibility. Statistical methods such as logistic regression and likelihood ratios~\cite{Lee2006,Regmi2014,reichenbach2018review}  can utilize a multitude of landslide controls as inputs. Scientists using these data-driven approaches have obtained a measurable degree of success in determining areas susceptible to landslides~\cite{Lee2006,Regmi2014,tien2019shallow}.  
However, these data-driven models also rely on the expert's choices, preconditions, and classifications of input variables. The outcome of these models' results, the landslide susceptibility map, does not decouple individual feature contributions to landslide susceptibility nor account for their interdependencies due to the limited computational capabilities in conventional approaches~\cite{reichenbach2018review}.

Machine learning approaches, such as fuzzy logic algorithms, support vector machines, and DNNs, have been applied to landslide studies for mapping landslide susceptibility~\cite{Pradhan2013,bui2020comparing,lsnn2}.   
DNNs have achieved improved 
performance compared to both statistical methods and other ML approaches
due to their use of nonlinearities, complex interdependencies of interlayer connections, as well as internal representations of data~\cite{lsnn1,lsnn2,bui2020comparing,van2020spatially,rudin2019stop,refx3,refx2}.   However, the black box 
nature of DNNs has been a major hurdle for their adoption in practice and research, making it difficult for experts to understand and trust their outcomes.  With DNNs, it is nearly impossible to determine the exact relation between individual inputs and outputs~\cite{rudin2019stop,refx3,refx2}.  Lack of interpretability is a weakness of DNNs and a fundamental drawback for high-stakes applications such as landslide mitigation where decisions impact lives and result in untold costs of insurance and reconstruction~\cite{cui2019cost,froude2018global,huang2013landslide}.  Interpretability would ideally provide decision-makers with a list of contributing factors ranked in order of importance, as well as any possible interplay between these factors.

The DNN’s lack of interpretability has prompted the Defense Advanced Research Projects Agency's (DARPA) third wave of AI call in 2017 and the European Union’s 2018 General Data Protection Regulation, which grants a right to an explanation, 
for algorithmic decisions that are made~\cite{European2020}.  Next-generation AI systems refer to the so-called explainable or interpretable AI (XAI) models.  The latter must be able to construct explanatory models for classes of real-world phenomena that can be communicated to humans~\cite{refx2}.  Various XAI categories have since been defined in the literature based on factors such as application and methodology, where each category is further divided into subclasses~\cite{refx1}.  Although the use of XAI in research is expanding, existing approaches aimed at explaining black box models exhibit a trade-off between accuracy and interpretability, resulting in a large gap in performance (e.g., \cite{leiva2019novel}). Recently, Rudin~\cite{rudin2019stop} showed that with proper feature engineering, and a shift from explaining existing black box models to creating methods with inherently interpretable models, the trade-off between accuracy and interpretability can be circumvented. 

To this end, we propose a 
framework that bridges the gap between explainability and accuracy for landslide susceptibility models. This framework utilizes a hybrid of model extraction methods and feature-based methods to generate a fully interpretable additive ANN model while simultaneously pruning features and feature interdependencies that are redundant or suboptimal to model performance and generalizability. Additive ANN are a 
type of generalized additive models (GAM) that have been recently gaining popularity~\cite{hastie1990generalized,hastie2017generalized,friedman2001greedy,agarwal2020neural}. They combine separate ANNs, each specializing in a single feature, to optimize a common outcome. Unlike other additive XAI methods such as Shapley additive explanations (SHAP) that aim to explain the local behavior of a black box model~\cite{lundberg2017unified}, additive neural networks are inherently interpretable models with both local and global interpretability. Model extraction methods aim to train an explainable ``student'' model to mimic the behavior of a ``teacher'' model, and feature-based methods aim to analyze and quantify the influence or the importance of each input feature~\cite{refx1}.  Our 
optimization framework possesses full interpretability, high accuracy, high generalizability, and low model complexity. Most notably, toy problems included in the Supplementary Note 1 demonstrate the capability of our framework to generate fully interpretable additive ANNs with controlled complexity and accuracy that can match state-of-the-art DNNs, as well as find globally optimal unique solutions. 
Furthermore,  
we utilize 
dataset division and outcome interpretation techniques uniquely suitable for landslide susceptibility modeling applications with spatially dependent data structures. 
We refer to the 
approach as superposable neural network (SNN) optimization in reference to the automated way of incrementally generating the additive ANN model and determining the contributing features. Our approach is different from the more commonly followed approach of designing a fixed network architecture with a fixed set of manually selected input features where the entire network is jointly trained in an end-to-end fashion~\cite{agarwal2020neural}.

In this study, we model three different regions of the easternmost Himalaya using SNNs. For comparison, we include results from a physically-based slope stability model (SHALSTAB), two statistical methods (logistic regression and likelihood ratios), in addition to state-of-the-art DNN teacher models. Finally, we examine the SNN-determined relationship and relative importance of each feature's contribution to landslide susceptibility and discuss how information extracted from the SNN can provide insights into the physical controls of landslides in our studied regions.  Our results highlight underappreciated, important controls such as the product of slope and precipitation and hillslope aspects in the studied region.  Controls that consist of products of input features can help unveil the influences from feature interactions.


\noindent {\bf Superposable neural networks.}  SNNs are an additive ANN architecture that enforces no interconnections between inputs (Fig. \ref{fig:DNNSNN}). The lack of interconnections between features is the key to explainability. Unlike DNNs where interdependencies between features are embedded in layers of network connections, interdependencies in SNNs are explicitly created as a product function of more than one original input feature. We refer to these products as ''composite features'' (see Methods for details). Important interdependencies between features are automatically determined by isolating composite features contributing to the desired outcome. Contributing composite features are explicitly added as independent inputs to the model, while non-contributing composite features are discarded (see SNN training flow diagram in Fig.~\ref{fig:SNNdiagram} as well as Methods).  Furthermore, we label SNNs according to the highest level of composite features used in training the model, which refers to the maximum number of features allowed in multivariate interactions. For example, a Level-3 SNN can include Level-1, Level-2 and Level-3 composite features. Using composite features, SNNs can approximate any continuous function for inputs within a specific range as a polynomial expansion to any desired precision. This ability allows SNNs to retain a level of accuracy on par with state-of-the-art DNNs.

The SNN is represented mathematically by the function (Eq.~\ref{eq:1}):
\begin{align}\label{eq:1} 
  S_t(\{\chi_j\}) =\sum_{j} \Bigl(  \sum_{k} w_{j,k} e^{-(a_{j,k} \chi_{j}+b_{j,k})^{2} }+c_{j} \Bigr).
\end{align}
It contains only two hidden layers of neurons with radial basis activation functions in the first layer and linear activation functions in the second layer.  The choice of radial basis activation functions allows the user to minimize the number of neurons in the model, maximizing the efficiency of our method. Each input $\chi_j$ is exclusively connected to a group of neurons to form an independent function $S_{j} = \sum_{k} w_{j,k} e^{-(a_{j,k} \chi_{j}+b_{j,k})^{2} }+c_{j}$ and the SNN output $\mbox{$S_t$} =\sum_{j}S_{j}$ is the sum of all independent functions, where $j$ = 1 : number of features ($M$), $k$ = 1 : number of neurons per feature ($v$), and $\chi_j$ is the $j\textsuperscript{th}$ composite feature. In addition to determining the features and interdependencies between features that contribute to the outcome, the SNN architecture enables the quantification of their exact contributions to the output. 

The model simplicity and lack of connections between neurons associated with different features makes our model fully interpretable and mathematically analyzable. However, this aspect also makes the model highly constrained, which poses challenges on its training. Jointly training the model with commonly used gradient descent-based optimizers proved to be extremely difficult to converge, especially as the number of features increases. Our 
optimization approach enables the separate training of individual neural networks by utilizing several state-of-the-art ML techniques (multi stage training, knowledge distillation, second order optimization~\cite{khaled_r8,khaled_r9,khaled_r10,khaled_r6,kdref,tan2018distill}) to deliver a model that is optimal in terms of performance and remarkably simple in terms of architecture.  The reduction in model complexity, while maintaining an accuracy that rivals that of DNNs, which are orders of magnitude more complex in terms of number of parameters and redundancies in interconnectivities, presents a substantial advance.

A validation of our approach using toy models is included in Supplementary Note 1.1 and 1.2. 
In the first application, we create a synthetic dataset by adding known functions of composite features and test the ability of the SNN to find the contributing features and extract their functions from the data. The second application incorporates up to Level-4 feature interactions and demonstrates the impressive ability to extract boolean relationships from synthetic data. Boolean inference tasks are notoriously difficult because of the high degree of stiffness and nonlinearity between input and output. The SNN optimization algorithm is described in Methods.

\noindent {\bf Landslides in easternmost Himalaya.} 
Asia holds the majority of human losses due to landslides globally, with a high concentration in the Himalayan Arc~\cite{petley2012global,froude2018global}. In particular, the easternmost Himalaya has a high susceptibility to numerous landslides from steep slopes, extreme precipitation events, flooding, and frequent earthquakes~\cite{Larsen2012,BookhagenBurbank2010,Barros2004,yang2018atmospheric, Ben-Menahem1974} (Fig. \ref{fig:studyreg} and Supplementary Figure 1). We generated a landslide inventory of the easternmost Himalaya by combining the manual delineation of landslide areas with a semi-automatic detection algorithm~\cite{ghorbanzadeh2019evaluation,prakash2020mapping} (Fig. \ref{fig:R123_LS_060222}a-c; a flowchart diagram in Supplementary Figure 2, exemplary landslides in Supplementary Figure 3). Within the entire study area of 4.19$\times$10$^9$ m$^2$, the total number of mapped landslides is 2,289, and their areas range from 900 to 1.96$\times$10$^6$ m$^2$ (Supplementary Table 1, Supplementary Figure 4)~\cite{S6/D5QPUA_2023}. Landslide densities calculated over a 2.25 km$^2$ window are generally high in the range front (max 0.121) and low in the hinterland ($\sim$0.039). 

Within the easternmost Himalaya, we selected three regions (the Dibang, Lohit, and range front regions) with varying ranges of landslide controls to test the performance and application of the SNN model (Fig.~\ref{fig:studyreg}). Hereafter, we refer to Dibang, Lohit, and range front regions as the N-S, E-W, and NW-SE regions, respectively. Testing the SNN over these three regions with varying environmental conditions will allow us to examine the following: 1) whether the SNN can identify universal or distinctly different controls of landslides, and 2) whether SNN-determined functions of feature contributions to susceptibility, $S_j$, are similar or different across these three regions. We used 15 single features in the SNN model (Supplementary Figure 5, Supplementary Table 2). The 15 single features include aspect (\emph{Asp}), mean curvature (\emph{Curv}$_M$), planform curvature, profile curvature, total curvature, discharge, distance to channel (\emph{Dist}$_C$), distance to faults (\emph{Dist}$_F$), distance to the Main Frontal Thrust and suture zone (\emph{Dist}$_{\emph{MFT}}$), drainage area, elevation (\emph{Elev}), local relief (\emph{Relief}), mean annual precipitation (\emph{MAP}), number of extreme rainfall events (\emph{NEE}) and slope. The inclusion of these variables is based on previous studies that examined landslide controls in the Himalayan region~\cite{devkota2013,Regmi2014,mandal2018,chowdhuri2021}. The details of study area, landslide inventory, input data sources and calculation are presented in Methods.

\noindent {\bf\textsf{Results and Discussion}}
 
\noindent {\bf SNN Implementation.}  We modeled landslide susceptibility of the easternmost Himalaya using Level-1, 2 and 3 SNN models. We find that the Level-3 SNN is able to achieve over 99\% of the accuracy of the state-of-the-art teacher DNN, and the Level-2 SNN is able to achieve over 98\%. Given the small difference, we assume the explainability of the Level-2 SNN to be sufficient for our analysis. Due to the nature of this application, a special data partitioning method was devised to partition each region into roughly 70\% for training and 30\% for valiudation, which utilizes Pythagorean tiling to partition the regions in a spatially representative manner (Fig.~\ref{fig:PT2}) (see Methods for details).

A threshold value of $S_t$ is used as a binary classifier to predict landslides and compare them with observed landslides from our inventory. We selected a threshold susceptibility corresponding to the closest point to a perfect classifying model with 100\% true positive rate and 0\% false positive rate on a receiver operating characteristic (ROC) curve. Areas with $S_t$ greater and lower than this threshold are classified as landslide (\emph{ld}) and non-landslide (\emph{nld}) areas, respectively, in the model (Fig.~\ref{fig:R123_LS_060222}d-f). 

\noindent {\bf Comparison with traditional landslide susceptibility modeling.}  In addition to the comparison against the state-of-the-art DNN teacher model, we provide comparisons of Level-1 and Level-2 SNN performance to a number of traditional methods, all applied to the same regions and using the same inventory data. Comparison of different models on the same area is needed since model performance cannot be directly compared to model performance published in other papers, since those papers focused on different regions.

First, we investigated each of the 15 single features as individual classifiers for landslide occurrences. Second, we applied a physically-based slope stability model (SHALSTAB) for soil landslides~\cite{Montgomery1994,Dietrich2001,Moon2011} that couples infinite slope stability and steady-state hydrology for cohesionless material. Considering that most landslides in our inventory are soil landslides (Methods), SHALSTAB was assumed to be suitable for our analysis. We modified SHALSTAB and calculated a metric called the failure index (\emph{FI}), as the ratio of driving to resisting forces on a hillslope. \emph{FI} is equivalent to the inverse of the factor-of-safety, which represents the propensity for landslide occurrence. Third, we used two commonly used statistical models, logistic regression and likelihood ratios, to model landslide susceptibility~\cite{Lee2005,akgun2012comparison,reichenbach2018review}. Logistic regression (hereafter, $LogR$) is based on a multivariate regression between a binary response of landslide occurrence and a set of predicting features that are continuous, discrete, or a combination of both types~\cite{Lee2005}. Likelihood ratios ($LR$) are calculated as the ratio of the percentage of landslide pixels relative to total landslide pixels divided by the percentage of pixels relative to the total area within a specific range of feature values~\cite{Lee2005,akgun2012comparison}. Previous studies have quantified the ratio of the probability of landslide occurrences to the probability of non-occurrences or all-occurrences within a range of feature values and referred to it as the likelihood ratio, frequency ratio, or probability ratio~\cite{Lee2005,akgun2012comparison,reichenbach2018review}.  A ratio of 1, $>$1, or $<$ 1 indicates an average, above-average, or below-average likelihood of landslide occurrence, respectively, within the feature range compared to that of the study area. Landslide susceptibility for each pixel is calculated as the sum of the corresponding $LR$ from each feature’s value. A threshold value of modeled landslide susceptibility from $LogR$ and $LR$ can be used as a binary classifier to predict landslides following a similar procedure that we used for the SNN. 

We assessed model performance based on various metrics including area under the receiver operating characteristic curve (AUROC).  In addition, we calculated the statistical measures of accuracy, sensitivity (probability of detection, POD), specificity (probability of false detection, POFD), and POD-POFD. We also calculated the 95\% confidence interval of mean AUROC from the statistical and neural network model outputs based on a 10-fold cross validation. The 95\% confidence intervals of mean AUROC can be used to determine whether model performances are statistically different (model and method details in Supplementary Note 2).  

We show that the SNN model's performance is comparable to that of the teacher, second-order-optimized DNN, while providing a statistically significant improvement over commonly used physically-based and statistical models. AUROCs of Level-1 and Level-2 SNNs are 0.856 and 0.890, respectively, calculated as the averages from the three study regions. The value for each region is presented in Supplementary Table 3. The Level-2 SNNs captured over 98\% of the teacher model (MST) performance across all three study regions. The Level-2 SNN is optimal in the sense that it provides high accuracy (comparable to deep nets) and relatively simple model complexity (hereafter, SNN refers to Level-2 SNN). 

The SNN achieved {\raise.17ex\hbox{$\scriptstyle\sim$}}21\% average improvement in AUROC over the top performing single original features (i.e., \emph{MAP} or slope, AUROC = 0.737), {\raise.17ex\hbox{$\scriptstyle\sim$}}22\% over a physically-based model (SHALSTAB) (AUROC = 0.727), and {\raise.17ex\hbox{$\scriptstyle\sim$}}5-8\% over logistic regression (AUROC = 0.848) and likelihood ratios (AUROC = 0.823) in our three study regions. The 95\% confidence intervals of the mean AUROC of the SNN lie above and do not overlap with those of the statistical models  (Supplementary Table 4). In addition, the vast majority of other performance metrics such as accuracy, POD, POFD, and POD-POFD from the SNN are improved over these other methods as well (Supplementary Table 5).

\noindent {\bf SNN model explainability.} The SNN-determined independent functions $S_j$ show varying relationships between both features and feature interdependencies, and their absolute susceptibility contribution (Fig. \ref{fig:All_fvS_052422}).  $S\textsubscript{\emph{MAP*Slope}}$ and $S\textsubscript{\emph{NEE*Slope}}$ generally exhibit steep increases with feature value, followed by asymptotic behavior (Fig. \ref{fig:All_fvS_052422}a, d, g). These nonlinear relationships between landslide susceptibility and the product of slope and climatic features of \emph{MAP} and \emph{NEE} are similar in all three regions. In addition, $S\textsubscript{\emph{Asp}}$ shows a peak around $145^\circ$ to $180^\circ$, which indicates a preference for south-facing slopes, likely due to moisture from the Bay of Bengal~\cite{BookhagenBurbank2010} (Supplementary Figure 6, Supplementary Note 3). These functional relationships are similar to those deduced by the $LR$ statistical method that represent the likelihood of landslide occurrence. However, unlike $LR$, which assume the same, average likelihood ($LR=1$) for each feature, $S_j$ corresponding to $LR=1$ varies depending on a feature’s absolute, decoupled contribution to landslide susceptibility.

The SNN provides the exact contribution of each individual feature to the total susceptibility outcome, which allows us to quantify the relative importance of landslide controls in different localities and across varying spatial scales (Fig. \ref{fig:Global_PieBar_052322}d-f). Causal rankings of individual features that drive landslides can be obtained by calculating the susceptibility difference between \emph{ld} v.s. \emph{nld} pixels, $\Delta \bar{S}_{j}$, within a region of interest for each individual feature. This is demonstrated both globally (Fig. \ref{fig:Global_PieBar_052322}a-c), where the region of interest is the entire region of study, and locally (Fig. \ref{fig:R123_Rank_060222}a-c), where the region of study is divided into hundreds of smaller regions of interest, each consisting of a 2.25 km$^2$ window. For comparison, we also identified the primary controls of landslides and their relative contributions from the Level-1 SNN and weights determined by the logistic regression model (Supplementary Note 2, Supplementary Figure 7).  

Composite features involving topographic and climate features are identified as important landslide controls for our study area. Namely, the product of slope and \emph{NEE} or \emph{MAP}, \emph{Asp}, and the product of \emph{Asp} and \emph{Relief} tend to have large $\Delta \bar{S}_{j}$ across all three regions (Fig. \ref{fig:Global_PieBar_052322}a-c). In addition, those features are identified as locally important, primary features when analyzing using a  2.25 km$^2$ window throughout the area (Fig. \ref{fig:R123_Rank_060222}a-c). The primary features of \emph{MAP*Slope} and \emph{NEE*Slope} are consistent among our three study regions in the easternmost Himalaya, despite differences in the spatial distribution and magnitude of precipitation and proximity to a major fault with a history of earthquakes (Supplementary Figure 1). Although these composite features may not be the largest contributor for total susceptibility (Fig. \ref{fig:Global_PieBar_052322}d-f), they tend to have different contributions for \emph{ld} and \emph{nld} areas and lead to a large {$\Delta \bar{S}_{j}$ (Fig. \ref{fig:Global_PieBar_052322}a-c).

SNN-derived individual feature contributions are used to assess the relative importance between climate and slope features. The feature independence in the SNN additive architecture and the use of composite features allows us to isolate the effect of slope or climate in the model. (1) The exact marginal contribution is calculated for Level-2 features involving slope or climate (i.e., \emph{Asp}, \emph{NEE}, and \emph{MAP}). (2) Level-1 slope and Level-2 slope marginal contributions are added together to produce the total susceptibility contribution from the slope, $S_{t,Slope}$. (3)  Level-1 climate and Level-2 climate marginal contributions are added together to produce total susceptibility contribution from climate features,  $S_{t,Climate}$. In Fig. \ref{fig:R123_Rank_060222}d-f, we compare the relative importance of slope and climate features using our approach that separates their contributions between \emph{ld} and \emph{nld} pixels throughout the region. Then, we calculate the difference between $\Delta \bar{S}_{t,Slope}$ and $\Delta \bar{S}_{t,Climate}$, divided by the threshold susceptibility value, $S_{t,threshold}$, for each respective region. We find that  {\raise.17ex\hbox{$\scriptstyle\sim$}}74\%, 54\%, and 54\% of localities have a larger contribution from climate features than that of slope for the N-S, NW-SE, and E-W regions, respectively, emphasizing an overall importance of climatic features that drive landslides.


\noindent {\bf Accurate and interpretable landslide susceptibility from the SNN.} Whereas many XAI efforts involve a trade-off between accuracy and interpretability, our 
SNN does not compromise accuracy. Given the SNN’s inherent and unique ability to decouple individual feature contributions and select feature interdependencies, we can easily isolate local contributions from primary controls discovered by the SNN (Fig. \ref{fig:R123_Rank_060222}). Our local analyses for assessing landslide controls indicate that the contribution of climate features, such as \emph{NEE}, \emph{MAP}, and \emph{Asp}, to landslide susceptibility tends to surpass that of slope for a majority of landslide occurrences in this area. 
These results highlight a prevalent climatic control on landslide occurrences in the easternmost Himalayan region. Due to the eastward increasing trends of precipitation rate and variability along the Himalaya, the easternmost Himalaya contains one of the largest strike-perpendicular climatic variations across the steep mountain range~\cite{BookhagenBurbank2010}.  This considerable climate gradient from the range front to the hinterland likely impacts landslide susceptibility in the easternmost Himalaya. 

The transparency of our SNN model offers insight into potential mechanisms of landslides and the relative importance of controlling factors. First, the SNN highlights the important, yet under-appreciated controls of {\emph{NEE*Slope}}, {\emph{MAP*Slope}}, {\emph{Asp}}, and {\emph{Asp*Relief}} (Fig. \ref{fig:R123_Rank_060222}), which implies a dominant occurrence of precipitation-induced landslides in our study site. However, these topography-climate composite features reveal the importance of both incorporated features.  These features comprising the product between slope and precipitation rates and intensity as well as that of aspect and relief suggest that landslides are affected by strong slope-climate couplings and aspect-related microclimates. 

The nonlinear asymptotic function of $S\textsubscript{\emph{MAP*Slope}}$ and $S\textsubscript{\emph{NEE*Slope}}$ (Fig. \ref{fig:All_fvS_052422}a, d, g) can be explained by a physical mechanism of rainfall-induced landslides that induces slope failure due to an increase in pore-water pressure and subsurface saturation~\cite{iverson2000landslide}. The modeled total landslide susceptibility ($S_t$) is analogous to the physically-derived failure index (\emph{FI}), which is equivalent to the inverse of the factor-of-safety.  \emph{FI} is formulated from equilibrium on an infinite, cohesionless slope considering a pore pressure effect based on SHALSTAB ~\cite{Montgomery1994,Moon2011} as:
\begin{align}\label{eq:3} 
    FI = \frac{S}{S_0} \left( 1 - W \frac{\rho_w}{\rho_s} \right)^{-1}   
\end{align}
where $S_0$ is the threshold slope, $S$ is the local slope, $\rho_s$ is the wet bulk density of soil (2.0 g/cm$^3$), $\rho_w$ is the bulk density of water (1.0 g/cm$^3$), and $W$ is wetness. $W$ is calculated as a ratio between local hydraulic flux from a given steady-state precipitation rate relative to that of soil profile saturation~\cite{Montgomery1994}:
\begin{align}\label{eq:4} 
    W = \frac{h}{z} = \frac{qA}{bT\sin\theta}
\end{align}
where $h$ is the saturated height of the soil column ($L$), $z$ is the total height of the soil column ($L$), $q$ is the steady-state precipitation during a storm event ($L$/$T$), $A$ is the drainage area (\emph{L\textsuperscript{2}}) draining across the contour length $b$ ($L$), $T$ is the soil transmissivity when saturated ($L\textsuperscript{2}/T$), and $\theta$ is the local slope in degrees. $W$ varies from 0 (unsaturated) to 1 (fully saturated).  See Supplementary Note 2 for model details. 

Expansion of the denominator in a geometric series gives:
\begin{align}\label{eq:5} 
	FI = \frac{S}{S_0} \left( 1 + W \frac{\rho_w}{\rho_s} + W^2 (\frac{\rho_w}{\rho_s})^{2} + O(W^3) \right)	
	\equiv \frac{S}{S_0} k(W).
	\end{align}
The approximated \emph{FI} has three components: local slope $S$, threshold slope $S_0$, and \emph{k(W)}, which represents the degrees that landslides are promoted by subsurface saturation. \emph{k(W)} varies from 1 (unsaturated) to 2 (fully saturated). The multiplication of local slope and \emph{k(W)}, which has an upper bound, mimics the nonlinear asymptotic function of $S\textsubscript{\emph{MAP*Slope}}$ and $S\textsubscript{\emph{NEE*Slope}}$. This asymptotic increase in susceptibility is similar to observations of other precipitation-induced landslides, but different from earthquake-induced landslides whose occurrences increase nonlinearly with increasing slope~\cite{meunier2008topographic,huang2014topographic}. 

Second, the identified controls of {\emph{MAP}}, {\emph{NEE}}, and {\emph{Asp}} imply that local precipitation infiltration on steep slopes may be the dominant contributors to subsurface saturation in the easternmost Himalaya. A change in climatic conditions can raise volumetric water content and porewater pressure. This rise leads to an increased degree of subsurface saturation (i.e., $W$) and subsequently induces slope failure. Previous physically-based slope stability models consider various climatic factors (e.g., rainfall amount and intensity, subsurface convergence flow) to deduce the degree of subsurface saturation to model rainfall-induced landslide occurrences~\cite{Montgomery1994,Baum2002,Baum2010}.  For example, SHALSTAB~\cite{Montgomery1994,Dietrich2001} uses the topographic wetness index, proposed by Beven and Kirkby (1979)~\cite{Beven1979}, to calculate subsurface saturation considering the convergence of shallow subsurface flow from up-slope drainage areas for a given steady-state precipitation. On the other hand, the Transient Rainfall Infiltration and Grid based Regional Slope stability model (TRIGRS)~\cite{Baum2002, Baum2010} calculates transient pore pressure development due to vertical rainfall infiltration from rainfall intensity. In reality, both subsurface convergence and rainfall infiltration are essential contributors to subsurface saturation and need to be implemented in physically-based slope stability models. However, measuring precipitation intensity, moisture availability, or subsurface convergence and saturation in the field is difficult, especially in rural mountainous areas with limited accessibility. 

According to our SNN model results, the most important, controlling features for landslides in this area are the product of slope and \emph{MAP} (N-S region) or that of slope and \emph{NEE} (NW-SE and E-W regions). This result implies that local precipitation infiltration influenced by precipitation rate and intensity, represented by \emph{MAP} and \emph{NEE}, may serve as a first-order control on $W$ or \emph{k(W)} in eq.~(\ref{eq:5}). The absence of drainage area or discharge as a dominant contributing feature to susceptibility may suggest that subsurface flow convergence may be a second-order contributor to landslides in the easternmost Himalaya. However, we cannot rule out the possibility that the importance of topographic convergence was masked due to the low-resolution of our input topographic and rainfall data~\cite{leonarduzzi2021numerical}. These factors can be further examined in future studies using high-resolution topographic and climate data in SNN models.

Nonetheless, identifying the exact trigger for a landslide requires dense field measurements and historic records of soil, hydrologic, and climatic conditions (e.g., soil moisture, antecedent rainfall, rainfall intensity)~\cite{Orland2020,kirschbaum2020changes}, which are often difficult to obtain, especially in rural mountainous areas with limited accessibility. We have shown that our SNN model can identify key controls and quantify their potential contributions to susceptibility, highlighting the essence of strong slope-climate coupled controls on landslide occurrences. The composite features identified by the SNN such as \emph{NEE*Slope} or \emph{MAP*Slope} are consistent with previous understandings of landslide mechanisms. However, they were not explicitly implemented in previous data-driven statistical models. In DNNs, such couplings would likely be identified, but if that were the case, the information would be implicitly contained in the network weights and not readily available to the user. By incorporating climatic composite features including \emph{MAP*Slope}, \emph{NEE*Slope}, and \emph{Asp*Relief}, the performance of the SNN improved, increasing average AUROC by 5-22\% 
compared to those of statistical or physically-based models~\cite{Montgomery1994,Dietrich2001,Lee2005,akgun2012comparison} (Supplementary Note 2, Supplementary Table 3).  This performance enhancement is statistically significant according to our confidence interval estimates from a 10-fold cross validation.

\noindent {\bf Implications, limitations, and future directions.} Our work presents a substantial advance in XAI applications to natural hazards and circumvents the “black box” nature of common AI models. SNNs provide quantitative analyses of controlling factors and further highlight the important, mechanistic interpretations of landslides. Our 
AI-based decision-making approach provides a comprehensive framework that allows for the examination of numerous composite features and identification of key controls while retaining high accuracy. As natural perturbations increase due to urban development and climate change, the SNN may provide a promising, data-driven predictive tool that will enable communities to confidently tailor plans for hazard mitigation. 

While a variety of explainable AI methods are available today, our proposed SNN method offers unique advantages that are not simultaneously present in any other method. SNN is a fully explainable model that achieves a level of explainability comparable to linear regression, while delivering state-of-the-art performance that matches that of black box models like deep neural networks. Furthermore, unlike other additive models, SNN can incorporate multivariate functions without compromising full explainability. Additionally, the model features adaptive optimization of both feature selection and network architecture during training. A comprehensive comparison of SNN with other explainable AI methods must take all of these factors into account. This requires an in-depth study beyond the scope of this paper. For instance, other additive model methods generally rely on fixed architectures and preselected feature sets that lack feature interactions beyond bivariate interactions. On the other hand, decision trees utilize highly nonlinear interactions between multiple features through a different approach that theoretically offers full explainability, but is often difficult to interpret for large number of features or complex problems requiring numerous branches. It is also worth noting that SNN is not restricted to MST as the teacher model, and its accuracy can be further improved when more accurate teacher models are found. A viable alternative to MST for applications with small datasets is random forest, which is an ensemble of decision trees trained on randomly selected feature and dataset subsets using bootstrapping. While decision trees are explainable, random forest is considered a black box since its outcome is an aggregate of multiple trees. In such cases, SNN can leverage random forest as a teacher model to achieve similar accuracy while maintaining full explainability.

We acknowledge that the overall importance of slope and climatic features and their functional relationships with susceptibility revealed by the SNN are qualitatively similar to those inferred from statistical models. However, the SNN is more useful for landslide susceptibility assessment because it decouples individual feature contributions and quantifies absolute contributions from features and feature interdependencies. For example, the relative and absolute importance of SNN decoupled features are different from those determined by the weights set by logistic regression. In addition, our analysis shows that $S_j$ corresponding to $LR=1$ differs depending on a feature’s absolute, decoupled contribution to landslide susceptibility. The SNN approach reveals the important coupling between slope and climatic factors (e.g., \emph{MAP*Slope}, \emph{NEE*Slope}) as a primary driver for landslide occurrence. Accounting for these under-appreciated features and feature interdependencies that are not generally implemented in statistical methods or physically-based models can lead to a substantial increase in performance.  We note that these results are specific to the region analyzed herein (easternmost Himalaya), and other regions may feature a different set of dominant factors. 

We acknowledge there are limitations of our method in the easternmost Himalaya. Our input features are averaged over time and space, making it impossible to relate them directly to specific events (e.g., intense rainstorms or earthquakes) inducing  landslides in our inventories. In addition, our inventory is based on optical satellite images acquired at a specific time (e.g., 2017 Landsat) and post-failure spectral signatures. Thus, our model lacks information about the precise timing or types of  landslides (e.g., fast- or slow-moving landslides, soil or bedrock landslides). This makes it difficult to assess the timescales and spatial dependencies of landslide-triggering events (e.g., rainfall intensity or duration) for specific landslides or landslide types. Previous studies from the Nepal Himalaya suggest that the spatial distribution of landslides can vary with triggering events such as cloud outbursts, flooding and large-magnitude earthquake~\cite{jones2021natcomm,jones2021jgr}. 

However, for this study region, our method properly captures the first-order climatic controls of landslide occurrences. Our primary feature datasets may capture a representative, spatial distribution of landslide-triggering events such as intense precipitation and rock damage over the decadal timescale of concern. In the easternmost Himalaya, both MAP and NEE from TRMM and APHRODITE datasets covering 12 and 50 years show similar southward increasing trends~\cite{BookhagenBurbank2010,Yatagai2012}. This spatial pattern likely emerges from the aggregation of intense precipitation events influenced by orographic precipitation~\cite{BookhagenBurbank2010}. In the 30 years prior to the mapped inventory, there were no earthquakes with a magnitude larger than M$_W$ 5.0 (Incorporated Research Institutions for Seismology, www.iris.edu), which can induce abundant landslides. In future studies, a time-series landslide inventory from multiple years and information on nonrepresentative or infrequent extreme events can be used to assess the spatial and temporal correspondence between triggering events and landslides~\cite{jones2021jgr}.

Additionally, landslide and input feature data have relatively coarse spatial resolutions and are based on limited temporal information (e.g., 30 m resolution Landsat satellite images from 2017~\cite{EarthExplorer}, 90 m resolution SRTM DEM~\cite{EarthExplorer}, and {\raise.17ex\hbox{$\scriptstyle\sim$}}5 km$^2$ resolution TRMM data over 12 years~\cite{BookhagenBurbank2010}). We do not have access to high-quality, high-resolution data of topography, surface materials (e.g., soil depth, bedrock structures, lithology), and climatic and ecohydrologic conditions (e.g., landslide-triggering storm intensity, time-series precipitation intensity, vegetation types). Due to the extremely rugged mountains in the Himalaya, the highest available DEM resolution without extensive data gaps, suitable for regional-scale landslide susceptibility analysis, is 90 m~\cite{kirschbaum2020changes,stanley2017heuristic}.  Also, there are no readily available time-series precipitation data with a resolution $<$5 km$^{2}$ in this area. We used relatively coarse 30 m resolution Landsat images to map landslides even though limited high-resolution satellite imagery is available (e.g., Planetscope Scene). This is because: 1) Landsat images are globally available, open-source satellite images with a {\raise.17ex\hbox{$\scriptstyle\sim$}}40-year historic archive, 2) reliable topographic, climatic, and geologic feature data have coarser resolutions than 30 m, and 3) we cover a large region of the easternmost Himalaya (a total area of 4.19$\times$10$^9$ m$^2$, 4.66$\times$10$^6$ pixels at 30 m). When applying a regional-scale model covering a large area with limited input data resolution and high computational costs, the use of 30 m resolution imagery for our model was inevitable. Although our inventory is based on coarse 30 m resolution Landsat images, our landslide inventory adequately captures the regional-scale spatial distributions of landslide occurrences and provides essential information for regional-scale landslide susceptibility models (see Methods). However, it is possible that our results 
from both physically-based or data-driven models may be biased due to the inherited uncertainties and limitations of our input data that are resolution-sensitive (e.g., topographic metrics, mapped landslides). 

Despite data limitations and uncertainties, our method is general and adaptable to other regions as well as sets and formats of contributing factors and available datasets.  Our SNN analysis of the easternmost Himalaya alone presents an important contribution to landslide hazard studies. High mountains in Asia hold the majority of human losses due to landslides globally, according to a global analysis conducted using 2004 - 2016 data~\cite{petley2012global,froude2018global}. Due to the associated high risks, there have been efforts to model landslide susceptibility in the Himalayan regions based on currently available data with limited resolutions~\cite{devkota2013,Regmi2014,mandal2018,chowdhuri2021,kirschbaum2020changes}.  Our work aims to capture the regional-scale spatial distributions of landslide susceptibility, differentiate controls of landslide occurrences, and provide interpretable, empirical functional relationships between landslide controls and susceptibility.
The decoupled SNN-identified functions combined with future changes in environmental conditions (e.g., extreme precipitation)~\cite{kirschbaum2020changes, stanley2020building} may provide a promising tool for assessing potential landslide hazards in this area. Additionally, a modified version of the semi-automatic detection algorithm can be extended further to incorporate InSAR data from sources such as Copernicus Sentinel-1 satellites alongside time-scale optical satellite imagery~\cite{bekaert2020insar,singh2022detecting} to specifically detect slow-moving landslides in future studies. With these datasets, we can apply SNN methods to slow-moving landslides and assess the controls of surface deformation while accounting for temporal changes in environmental conditions~\cite{finnegan2021unsaturated}. 
Our method is easily applicable to other locations, different datasets, and other physical hazards, such as earthquakes and wildfires.  The SNN is remarkably simple consisting of only two hidden layers, yet its performance rivals that of DNNs. Our SNN can also be easily updated and improved when global, open-source, high-resolution datasets and high-performance computational resources become more available in the future.



{\bf Data Availability:} The manual and semi-automatically mapped landslide datasets used within this manuscript are provided as polygon shapefiles through the UCLA Dataverse: \\
\texttt{ https://doi.org/10.25346/S6/D5QPUA }

\newpage
\section*{Methods}


\section*{Study Area}
Numerous landslides in the Himalayan region come from steep topography, intense rainfall and flood events, and seismic activities~\cite{Larsen2012,Coudurier2020,BookhagenBurbank2010,chowdhuri2021,Kent2004}. In particular, the easternmost Himalaya (Fig. \ref{fig:studyreg}) has a high susceptibility to landslides due to the following reasons. First, this area exhibits a dramatic precipitation gradient due to moisture originating from the Bay of Bengal in the south~\cite{BookhagenBurbank2010,Barros2004,yang2018atmospheric} (Fig.~\ref{fig:studyreg}). Previous studies have calculated daily and mean annual precipitation rates based on 90-min measurements from the Tropical Rainfall Measuring Mission (TRMM) 2B31 over 12 years (January 1998 to December 2009), with a spatial resolution of {\raise.17ex\hbox{$\scriptstyle\sim$}}5 km$^2$~\cite{BookhagenBurbank2010}. According to these datasets, our region has mean annual precipitation rates (\emph{MAP}) varying from {\raise.17ex\hbox{$\scriptstyle\sim$}}7000 mm/yr in the range front to {\raise.17ex\hbox{$\scriptstyle\sim$}}200 mm/yr in the hinterland~\cite{BookhagenBurbank2010} with the number of extreme rainfall events (\emph{NEE}), calculated as the number of days that exceed the 90\textsuperscript{th} percentile of daily rainfall rates,  reaching {\raise.17ex\hbox{$\scriptstyle\sim$}}13 and {\raise.17ex\hbox{$\scriptstyle\sim$}}2 events/yr in the range front and hinterland, respectively~\cite{BookhagenBurbank2010}.  The dramatic orographic patterns of precipitation magnitude and variability are also observed in the 57-yr Asian Precipitation–Highly Resolved Observational Data Integration Towards Evaluation of Water Resources project (APHRODITE)~\cite{Yatagai2012}. Second, this area has consistently steep slopes from the range front, where Holocene Himalayan shortening is concentrated near and along the Main Frontal Thrust, into the hinterland, which is affected by deglaciations from the last glacial maximum~\cite{Burgess2012,Haproff2019,Haproff2020,Salvi2020}. Third, this area is prone to active seismicity. The 1950 M$_W$ 8.6 Assam earthquake, one of the largest earthquakes in the Himalayan range, struck the nearby Namche Barwa region~\cite{Ben-Menahem1974}.  Since 1973,  this region has experienced $>$450 earthquakes with M$_W$ $>$4 according to the Incorporated Research Institutions for Seismology data archive (www.iris.edu, accessed on 10/01/2020). Many of these factors contribute to landslide occurrences in our study site. 

Within the easternmost Himalaya, we selected three regions (the Dibang, Lohit, and range front regions) with varying ranges of landslide controls to test the performance and application of the SNN model (Figs.~\ref{fig:studyreg} and Supplementary Figure 1). Both Dibang and Lohit regions extend from the active range front to the hinterland, from north to south and east to west, respectively. The Dibang region consists of metasedimentary rocks in the range front and crystalline rocks in the hinterland. The Lohit region is mainly composed of crystalline rocks. The active range front region is oriented in a northwest-southeast direction and mainly composed of metasedimentary rocks.

\section*{Landslide Inventory}

We generated a landslide inventory of the easternmost Himalaya using a semi-automatic detection algorithm that combines manual delineation of landslide areas with an automatic detection algorithm based on convolutional neural networks (CNN)~\cite{ghorbanzadeh2019evaluation,prakash2020mapping,S6/D5QPUA_2023} (Fig. \ref{fig:R123_LS_060222}a-c; the method illustrated using a flowchart diagram in Supplementary Figure 2). The basic procedure is as follows. We initially mapped landslides using 30 m resolution Landsat 8 imagery from November 2017 with bands 2, 3, 4, 5, and 7~\cite{EarthExplorer}. These satellite images were used to generate natural and false color imagery to show information of landcover types. High degrees of vegetation in the area allow for the easy detection of vegetation removal due to landslides and clear delineation of a landslide polygon. Most landslides are mapped as a combination of source and deposit, which are difficult to distinguish in coarse resolution Landsat bands. Whenever possible, we excluded debris transport or deposits and only mapped landslide scars associated with source areas. Because our landslide mapping is based on spectral signatures of post failures, our inventory likely includes both shallow, soil landslides and deep, bedrock landslides. 

We only assessed regions where landslides generally have the potential to occur or be detectable. Thus, areas of topographic slope less than 0.06 and alpine areas without vegetation cover were excluded from our landslide mapping and analysis. A slope threshold of 0.06 was determined to be the minimum slope along which landslides occur based on a cumulative distribution function of slope from observed landslides in the easternmost Himalaya. Similar criteria based on terrain characteristics such as slope or local relief have been used in previous studies to constrain the area of landslide analysis~\cite{parker2011mass}.  Alpine areas were classified using spectral signatures representing snow cover in Landsat 8 imagery from February 2018. 

Then, we used a CNN to detect landslides automatically, following previous works~\cite{ghorbanzadeh2019evaluation,prakash2020mapping} (Supplementary Figure 2). The CNN is used as a segmentation model for identifying landslides from 5 Landsat 8 bands and 7 input features (i.e., mean curvature, elevation, local relief, mean annual precipitation, slope, failure index, and wetness). The model takes a 32$\times$32$\times$12 patch as an input, where 12 represents the sum of 5 satellite bands and 7 input features. The model produces a 32$\times$32 binary patch as an output, where landslide pixels are given a value of 1, and non-landslide pixels are given a value of 0. The model segments a full region by dividing the region into 32$\times$32 patches, segmenting each patch individually, then stitching the model outputs back together to obtain a fully segmented region. The training dataset was prepared by manually annotating a small percentage of each studied region to be used as the ground truth targets for training the CNN. The manually annotated areas were selected as a number of randomly distributed 50$\times$50 pixel square sections throughout the studied regions. The manually annotated sections were selected such that half of them include landslides and half of them do not.  Hundreds of 32$\times$32 patches were extracted from each 50$\times$50 square section to augment the size of the training dataset. Once the CNN model is trained and used to segment the full region, the result is reviewed manually by an expert and modifications are made.

We manually corrected landslides from the automatic detection method using Landsat 8 images, high-resolution satellite images from Google Earth, and a 4-band Planetscope Scene with a 3 m resolution. Manual correction is necessary because of potentially inaccurate representations of landslide areas in automatically mapped inventories. Common issues include large detected features aggregated from multiple, adjacent landslides and small detected features that are not related to landslides~\cite{marc2015amalgamation,parker2011mass}.  We divided aggregated features into multiple landslides following suggestions from a previous study~\cite{marc2015amalgamation}. Most landslide polygons in all study regions were checked for aggregated features, which were divided based on the spectral signatures of recent scars and debris flows shown in high-resolution imagery. We used the manually corrected, automatically mapped landslides for our final landslide inventory (referred to as semi-automated landslides)~\cite{S6/D5QPUA_2023}. The spatial distributions and extents of landslides from our inventory are shown in Fig.~\ref{fig:R123_LS_060222}a-c.

The manually and semi-automatically detected landslides show a good correspondence [$>$90\% match for landslides $>$4 pixels (3,600~m$^2$)] based on object identification that examines the existence of overlapping areas. Generally, most landslides missing from the manually detected inventory are objects with a small number of pixels that are not easily and objectively detected by humans. Semi-automated landslides with $\leq$4 pixels comprise $\sim$7.5\% of total landslide areas. When comparing these pixels with 3 m resolution Planetscope Scene satellite images during the post-processing procedure, we found that many of these pixels are indeed small landslides showing different spectral signatures (e.g., Supplementary Figure 3). Thus, we included these semi-automatic landslides with $\leq$4 pixels in our final inventory. Areas commissioned by semi-automatic detection, but not manual mapping, were $\sim$0.1, $\sim$0.4, and $\sim$0.1\%, while areas omitted by semi-automated detection were $\sim$0.2, $\sim$0.6, and $\sim$0.1\% of the N-S, NW-SE, and E-W study areas, respectively.  

The area frequency distribution of our landslides from manual and semi-automatic mappings before 2017 shows a similar distribution to that of pre-2007 landslides from a nearby eastern Himalayan region that were manually mapped using 15-30~m resolution ASTER and Landsat images~\cite{Larsen2010,Larsen2012} (Supplementary Figure 4). According to a global compilation of geometrical measurements and types of 4,231 landslides~\cite{Larsen2010}, soil landslides from all examined regions including the Himalayan region do not appear to exceed an area of 100,000 m$^2$. Below this threshold, soil landslides tend to be dominant~\cite{Larsen2010,Larsen2012}. In our landslide inventory, $<$1\% of individual landslides and $<$20\% of total landslide area are greater than 100,000 m$^2$ (Supplementary Table 1). Thus, we assume that most mapped landslides are likely soil landslides.  In addition, we find that more abundant small landslides detected using the semi-automated method are similar to those observed in the landslide area-frequency distribution based on high resolution imagery ({\raise.17ex\hbox{$\scriptstyle\sim$}}4-15 m) from an eastern Himalayan region nearby (Supplementary Figure 4)~\cite{Larsen2012}. This supports that our semi-automatically mapped landslide inventory likely includes many small landslides missed by humans that were detected by a CNN-based automatic detection algorithm.

The total number of semi-automatically mapped landslides in our inventory is 2,289, whose areas range from 900 to 1.96$\times$10$^6$ m$^2$ (Fig.~\ref{fig:R123_LS_060222}a-c). The total mapped landslide area is 2.83$\times$10$^7$ m$^2$, which produces a landslide density of 0.007 within the entire study area of 4.19$\times$10$^9$ m$^2$ (Supplementary Table 1). Landslide density is also calculated within a 2.25 km$^2$ window, which is greater than the largest landslide size (1.96 km$^2$). Landslide densities calculated over a 2.25 km$^2$ window are high in the range front (maximum of 0.121) and low in the hinterland (maximum of 0.039).


\section*{Model Input Feature Descriptions}  
We quantified the spatial distribution of 15 topographic, climatic, and geologic controls and used them as input features for the SNN (Supplementary Figure 5, Supplementary Table 2). Topographic controls include aspect (the direction of topographic slope face; \emph{Asp}), mean curvature (\emph{Curv}$_M$), planform curvature, profile curvature, total curvature, distance to channel (\emph{Dist}$_C$), drainage area, elevation (\emph{Elev}), local relief calculated as an elevation range within a 2.5 km radius circular window  (\emph{Relief}), and slope. Climatic or hydrologic controls include discharge, mean annual precipitation (\emph{MAP}), and number of extreme rainfall events (\emph{NEE}). Last, geologic controls include the distance to lithologic boundaries (i.e., mostly faults) (\emph{Dist}$_F$) and distance to the Main Frontal Thrust and suture zone (\emph{Dist}$_{\emph{MFT}}$). These features were selected from literatures that examined landslide occurrences in the Himalayan region~\cite{devkota2013,Regmi2014,mandal2018,chowdhuri2021}. We mostly used features directly measured through satellite data including a 90 m digital elevation model from the Shuttle Radar Topography Mission (SRTM)~\cite{EarthExplorer} and rainfall magnitude and variability from TRMM~\cite{BookhagenBurbank2010}, as well as published regional geologic maps~\cite{Haproff2019,Taylor2009}. Utilizing open-source satellite data with a long-term historic archive allows anyone to easily implement our approach in other regions (e.g., Himalayan Arc) with limited accessibility, high landslide potential, and a long landslide history~\cite{zhu2019benefits,petley2012global,froude2018global,kirschbaum2020changes}. 

Below are the details of our data sources and methods of calculation. First, topographic variables such as slope, aspect, local relief, curvature, distance to channel, and drainage area were calculated from a 90 m SRTM digital elevation model (DEM)~\cite{EarthExplorer}.  Although a higher-resolution 30 m DEM is available, it contains missing values within our study area. Thus, we used a 90 m DEM for calculating topographic variables. Slope was calculated as the steepest descent gradient using an 8-direction (D8) flow routing method~\cite{Schwanghart2014}.  We calculated aspect, the direction of slope face, as the angle in degrees clockwise from north given by the components of the 3-D surface normal. The surface normal was calculated using the $x$, $y$, and $z$ components of each pixel. Local relief was calculated as the range in elevation within a 2.5~km radius circular window. We used a 2.5 km radius window because it is similar to the length scale of across-valley widths in the range front where most landslides are. Local relief at this scale allowed us to quantify the spatial variation of topographic relief relevant to landslides on these fluvial valleys. Curvature was calculated as the second derivative of the 90~m SRTM DEM. We calculated mean, planform, profile, and total curvatures using TopoToolbox 2~\cite{Schwanghart2014,Schmidt2003}.

To calculate distance from channel, we first determined flow direction using D8 flow routing. The flow direction was carved through topographic depressions and flat areas to avoid sinks and generate a continuous drainage system. We then imposed a minimum drainage area of 1 km$^2$ needed to initiate a stream before extracting a stream network based on the flow direction. Using the stream network, we calculated the distance of each pixel in the DEM to the nearest location in the stream network.

We acquired \emph{MAP} and \emph{NEE} from a previous study~\cite{BookhagenBurbank2010} that analyzed the Tropical Rainfall Measuring Mission (TRMM) 2B31 datasets from January 1998 to December 2009. Daily rainfall and \emph{MAP} values were integrated from 90-min measurements over 12~years. To calculate \emph{NEE}, the 90\textsuperscript{th} percentile of daily rainfall total for each pixel was determined for the 12-year measurement period~\cite{BookhagenBurbank2010}.  Only days with measured rainfall were included in calculating the probability density function. The number of days per year with a daily rainfall total above the 90\textsuperscript{th} percentile was counted as \emph{NEE}~\cite{BookhagenBurbank2010,Bookhagen2010}.  The resolution of the original \emph{MAP} and \emph{NEE} datasets in our study area is $\sim$5 km$^{2}$, which we resampled to 30 m resolution to be consistent with the resolution of our landslide inventory. To calculate the drainage area, we first calculated D8 flow directions of stream networks and calculated the number of upstream cells that contribute to each pixel. The number of cells can then be converted into a drainage area. Discharge was calculated by summing upstream contributing cells weighted by their \emph{MAP} to account for spatially varying precipitation patterns. Using these weights, cells with higher \emph{MAP} values will contribute more to total discharge than cells with lower precipitation values.

Previous studies~\cite{parker2011mass,xu2014three} have shown that distance to fault ruptures is a good predictor for the occurrence of earthquake-induced landslides. We do not have information on active fault planes at depth and ground peak acceleration patterns for past earthquakes in these regions. Thus, we calculated \emph{Dist}$_{\emph{MFT}}$ for our study regions as each pixel’s Euclidean distance from the closest point on traces of the Main Frontal Thrust (MFT) and suture zones mapped by Taylor and Yin~\cite{Taylor2009}. These faults represent potentially active faults in our study area~\cite{Haproff2019,Haproff2020}.  Because the suture zone is located far to the north, \emph{Dist}$_{\emph{MFT}}$ largely reflects the distance to the MFT. In addition, we calculated \emph{Dist}$_F$ as the Euclidean distance of each pixel from boundaries separating all lithologic units reported in~\cite{Haproff2019}. We included \emph{Dist}$_F$ because bedrock tends to be more damaged near major lithologic boundaries due to faulting, which may influence landslide occurrences. The Euclidean distance was calculated using ArcGIS 10.6.

\section*{SNN training method: composite features} 
We categorize composite features by the number of product operations involved. For example, given a problem with $n$ original input features $x_1, x_2,...x_n$, we can generate a set of $M \ge n$ composite features $\chi_1, \chi_2, ...\chi_M$, where Level-1 features are the single original features (first-degree monomials such as $x_i$) and Level-2 features are composite features equal to the product of two Level-1 features. As an example, we may form the product $x_1*x_2$ (second-degree monomial), where the monomials $x_1$ and $x_2$ are Level-1 features. Level-3 features are composite features consisting of a product of three Level-1 features, such as $x_1*x_2*x_3$ or $x_1*x_2^2$, and so on, resulting in third-degree monomials. Composite features are restricted to functions that cannot be derived from another function by elementary algebraic transformations. For example, $x_1^2 * x_2^2$ and $2 * x_1 * x_2$ are not permitted since they can be derived from $x_1 * x_2$ by elementary operations (namely, by squaring and scaling, respectively). In mathematics, composite features differing from each other by a finite number of elementary operations could define an equivalence class.

\section*{SNN training method: optimization} 
The flow diagram of the superposable neural networks (SNN) training
method is presented in Fig. \ref{fig:SNNdiagram}. The SNN is an
additive model~\cite{adnet1,adnet2} with a unique architecture
described by eq.~(\ref{eq:1}) and Fig. \ref{fig:DNNSNN}, and a unique training
method explained here. 

The method can be summarized by the following steps:
\begin{enumerate}
\item Multivariate polynomial expansion: composite features are generated.
\item Tournament ranking: an automated feature selection method we have
designed for finding the features that are most relevant to the model.
\item Multistage training (MST): a second-order deep learning technique for
generating a high-performance teacher network.
\item Fractional knowledge distillation: a technique we designed for
separating the contribution of each feature to the final output.
\item Parallel knowledge distillation: standard knowledge distillation
individually applied to networks corresponding to each feature.
\item Network superposition: merging single layer networks
corresponding to each feature into one SNN.
\end{enumerate}
The two stages of knowledge distillation are key in facilitating the optimization of the highly constrained SNN architecture in a way that maximizes accuracy while minimizing the number of neurons for optimal model simplicity. The multi stage training (MST) DNN used as the teacher model due to its high performance and regularization properties, was tuned to minimize the difference between training and testing accuracy to guide the SNN model into a regularized solution that avoids over-fitting. The steps are further explained in detail below.

\section*{SNN training method: multivariate polynomial expansion}  
Given $n$ features $x_1,x_2, \dots, x_n$, we generate $M$ composite features
$\chi_1,\chi_2,\dots ,\chi_M$ according to a predetermined maximum
composite feature level.

Ex. 1: If the original number of features is 3 and the maximum composite
feature level is Level-3, then we generate 13 composite features
$[\chi_1,\chi_2, \dots ,\chi_{13} ] = [x_1,x_2,x_3,x_1 * x_2, x_1 * x_3,x_2 *
x_3, x_1 * x_2 * x_3,  x_1^2*x_2, x_1^2*x_3, x_2^2*x_1, x_2^2*x_3,
x_3^2*x_1, x_3^2*x_2 ]$.

In this work, we have used 15 original features with a maximum composite feature Level-2. Because Level-3 performs marginally better than Level-2, we consider the Level-2 SNN as our optimal SNN. With 15 original features and the maximum composite feature Level-2, we generate a total 120 composite features. All features are standardized with zero-mean and unit-variance. The Level-1 SNN inputs are single features, and the Level-2 SNN inputs are single and composite features. The SNN output is the estimated total landslide susceptibility ($S_t$) at a specific location, which is the sum of the susceptibility contributions from all individual features. Our optimization approach allows for the exploration of multiple combinations of parameters (e.g., 120 composite features for Level-2) without relying on an expert's choices, preconditions, or classifications of input features. The initial set of potentially relevant features is determined by the tournament ranking step. The most relevant features are then iteratively determined during the training process, where the contribution of each control to susceptibility ($S_j$, where $j$ corresponds to a single or composite feature) is quantified using multiple steps of knowledge distillation. By superposing $S_j$, we produced (pixel-by-pixel) the total landslide susceptibility map, $S_t$, with values ranging from 0 to 1 as the final product (Fig.~\ref{fig:SNNdiagram}).

\section*{SNN training method: tournament ranking} 
Our feature selection technique is based on a point system and uses a combination of backwards elimination and forward selection~\cite{fsref} as building blocks. The composite features generated in the previous steps are randomly arranged into groups, with each group containing a subset of the features. Each feature group is used to train a simple neural network model. After the network is trained, backwards elimination is applied using area under the receiver operating characteristic curve (AUROC) as the performance criterion (Supporting Information). The top performing feature in the group receives a point. This process is repeated many times; several thousand groups were generated in the training of each SNN in this work. Features are ranked according to the points they accumulated. Forward selection is then applied in the order of the feature ranking to select the features that will be passed on to the next step.

The second-order Levenberg-Marquardt algorithm~\cite{khaled_r6} was used in training the individual neural networks models. It should be noted that using second-order training is essential for the practicality of this step. Unlike first-order training algorithms (based on gradient descent) that require manual hyper parameter tuning, second-order training algorithms are robust. In addition, second-order training can achieve better performance with fewer parameters~\cite{khaled_r1,khaled_r2,khaled_r3,khaled_r4,khaled_r5,khaled_r6,khaled_r7}. This allows for the automation of the process, and reduces the memory requirements for training the networks, yielding a more efficient parallel implementation on multicore processors.

\section*{SNN training method: multistage training} 
The high-ranked features that are passed on from the previous step are used to train a high-performance DNN. We chose MST as our DNN model, since it has shown superior performance in similar applications as well as regularization properties that counteracts over-fitting ~\cite{khaled_r8,khaled_r9,khaled_r10}.

\section*{SNN training method: fractional knowledge distillation}  
Knowledge distillation is a technique to reduce model complexity, by using the soft output of a more complex teacher DNN as the target of a less complex student DNN~\cite{kdref}. The MST in the previous step acts as our teacher network.

We have designed a variation of knowledge distillation that allows us to isolate the contribution of each feature to the estimated output. We call this variation fractional knowledge distillation, a term that is inspired by the fractional distillation technique in chemistry. We illustrate this using a step-by-step example for the case of two features. This can be easily generalized to any number of features.

Ex. 2: Assume that two composite features $[\chi_1,\chi_2]$ are passed on from the feature selection stage, and ordered according to importance where $\chi_1$ is the most important. Let $ts_0$ be the set of soft
targets obtained from the MST output:
\begin{enumerate}
\item Save a copy of $ts_0$, named $ts_{0c}$
\item Train a simple DNN $net_{1,1}$ using only $\chi_1$ as input and
$ts_0$  as an output
\item Obtain $o_{1,1}$, the set of outputs of $net_{1,1}$
\item Update $ts_0$ to $ts_0 - o_{1,1}$
\item Train a simple DNN $net_{2,1}$ using only $\chi_2$ as input and
$ts_0$ as an output
\item Obtain $o_{2,1}$, the set of outputs of $net_{2,1}$
\item Update $ts_0$ to $ts_0 - o_{2,1}$
\item Evaluate performance by calculating AUROC using 
$\sum_{i=1}^2\sum_{j=1}^1 o_{i,j}$  
and $ts_{0c}$
\item Train a simple DNN $net_{1,2}$ using only $\chi_1$ as input and $ts_0$ as an output
\item Obtain $o_{1,2}$, the set of outputs of $net_{1,2}$
\item Update $ts_0$ to $ts_0 - o_{1,2}$
\item Train a simple DNN $net_{2,2}$ using only $\chi_2$ as input and $ts_0$ as an output
\item Obtain $o_{2,2}$, the set of outputs of $net_{2,2}$
\item Update $ts_0$ to $ts_0 - o_{2,2}$
\item Evaluate performance by calculating AUROC using 
$\sum_{i=1}^2 \sum_{j=2}^2 o_{i,j}$ and $ts_{0c}$ 
\item Repeat $n$ times until the performance stops improving
\end{enumerate}

Each DNN above consists of only a few neurons and is trained for a small number of epochs where the contribution of each feature is gradually determined to avoid numerical instabilities. The number of neurons and epochs are hyper parameters that can be tuned based on the data.

\section*{SNN training method: parallel knowledge distillation} 
The outputs from groups of networks, corresponding to each feature from the previous step, are added together to yield one soft target per feature. Knowledge distillation is separately used to train a single SNN layer for each feature.

Ex. 3: Following the previous example:
\begin{enumerate}
\item Create two soft targets: $ts_1 = \sum_{j=1}^n o_{1,j}$, and $ts_2=\sum_{j=1}^n o_{2,j}$
\item Train a single layer network $net_1$ using $\chi_1$ as input and $ts_1$ as an output
\item Train a single layer network $net_2$ using $\chi_2$ as input and $ts_2$ as an output
\end{enumerate}

\section*{SNN training method: network superposition.} The single layer networks from the previous step are merged together to create the SNN, by adding an output layer that sums up the outputs of all the networks from the previous step. The connection weights at the output layer are set to one. The output of the SNN is a continuous value between 0 and 1, which  determines the network's estimation of landslide susceptibility at a specific location.

Ex. 4: Following the previous example, an SNN is created with $\chi_1$ and $\chi_2$ as inputs and $O=o_1 + o_2$ as the output, where $o_1$ is the output of  $net_1$ and $o_2$ is the output of  $net_2$.

\section*{SNN training method: implementation} 
In this work, we have created three SNNs for three regions. The data samples from each region were partitioned into roughly 70\% for training and 30\% for testing. All reported performance metric results in the paper were obtained using the testing portion of the data. Class imbalance was taken into consideration when training the networks. Given that the percentage of positive targets (locations containing a landslide) in each region is substantially smaller than negative targets (locations with no landslide), positive targets were weighted higher than negative targets in the training cost functions following the approach in Ref.~\cite{khaled_r10}.

\section*{Pythagorean Tiling}
While applying the SNN to landslide susceptibility modeling, we aimed to satisfy a number of conditions: (1) Full model interpretability, both locally and globally. (2) Minimizing the number of features included in the model. (3) Maximizing prediction accuracy. (4) Optimizing generalizability, such that the model is equally representative across each region. 

Due to the nature of this application, special attention should be paid to the last requirement. The standard practice in ML is to divide available data into two main partitions. One partition is used for training/validation (typically 70\% of the data) and the other one for testing (typically 30\% of the data). Traditionally, the goal is to maximize the reported accuracy of the testing partition where to a certain extent, over-fitting in the training portion of the data is not a primary concern. A key difference in this application is that a model generated for a certain region must be equally representative of and applicable to the entire region after training, both in accuracy and explainability. To meet this requirement, we use a special data partitioning technique that utilizes Pythagorean tiling to divide our data in a spatially representative manner that maintains variability between training and testing partitions. Using Pythagorean tiling, we generate a checkerboard pattern with a ~70/30\% square ratio, where bigger squares correspond to training and smaller squares correspond to testing (Fig.~\ref{fig:PT2}). Instead of primarily aiming to obtain the highest accuracy on the testing portion of the data, our algorithm is designed to find a more conservative solution with optimal balance between maximizing testing accuracy and minimizing the difference between training and testing accuracies.


\newpage

\begin{figure}[h] %
\includegraphics[width=\textwidth]{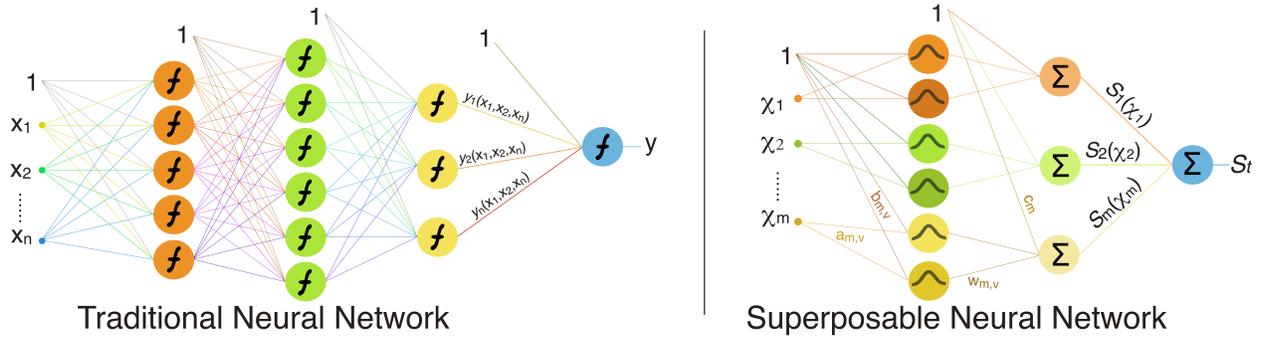}
\caption{\label{fig:DNNSNN}   {\bf  Conventional DNN architecture vs SNN architecture.} In a conventional DNN, features are interconnected and interdependencies are embedded in the network, making them virtually impossible to separate. In a SNN, features and feature interdependencies that contribute to the output are found in advance and explicitly added as independent inputs. Radial basis (Gaussian) activation functions are used in the SNN, where each neuron is connected to one input only. The $x_1, x_2,...x_n$ refer to a set of $n$ original features, and $\chi_1, \chi_2, ...\chi_M$ refer to a set of $M$ composite features. $y$ and $S_t$ refers to DNN and SNN outcomes of total susceptibility, respectively. The symbols in this figure are defined and explained in the main text, eq.~(\ref{eq:1}).}
\end{figure}

\newpage
\begin{figure}[h]
\includegraphics[width=0.7\textwidth]{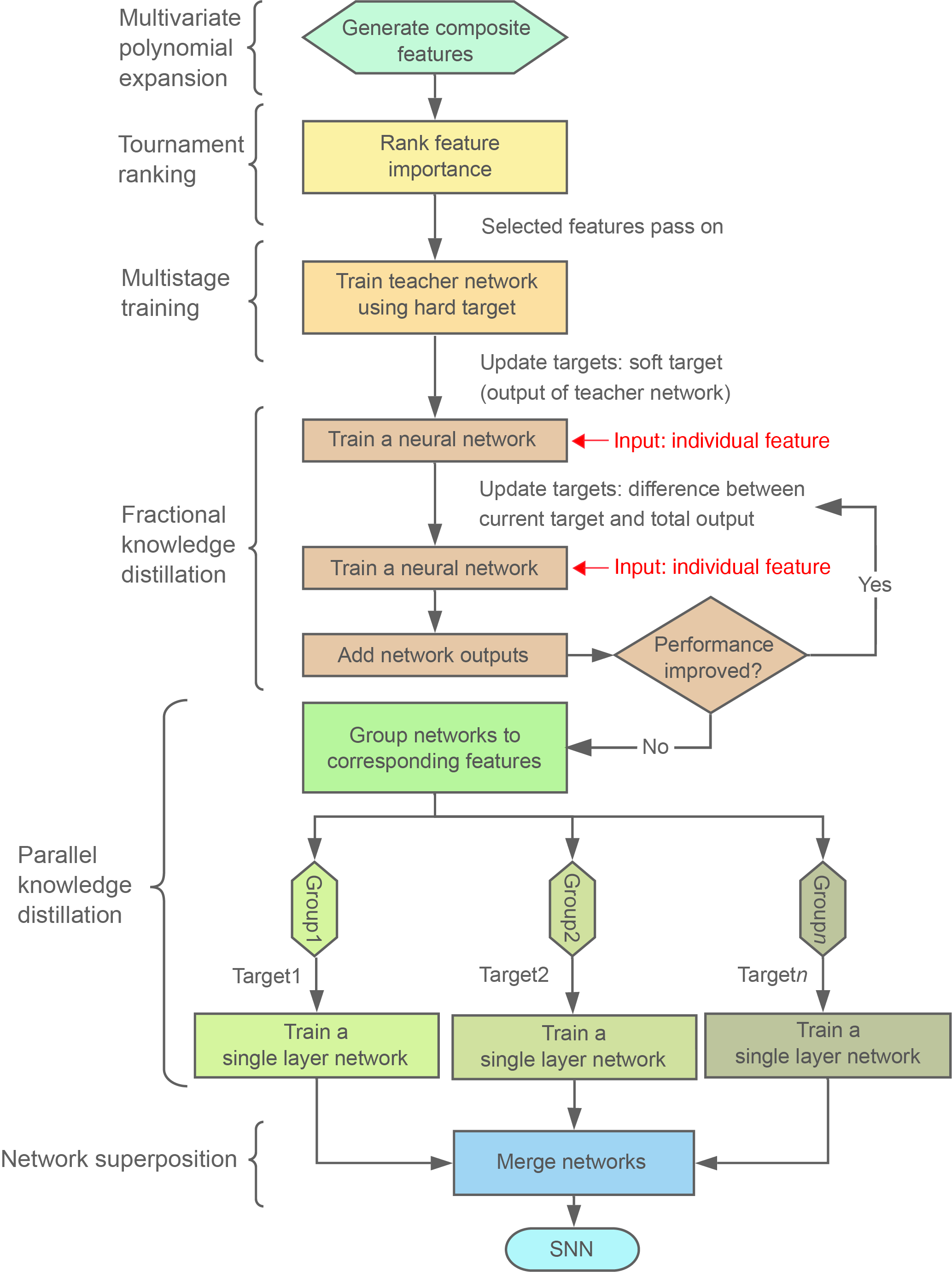}
\caption{\label{fig:SNNdiagram}  {\bf  Superposable neural network training flow diagram.} The flow diagram shows the methods used in our study, which include the feature-selection model and multistage training. Our feature-selection model based on multivariate polynomial expansion and tournament ranking allows for the exploration of multiple combinations of parameters without relying on an expert's choices, precondition, or classification of input features and identify a set of optimal composite features that are relevant to the landslide susceptibility. Then, multiple steps of knowledge distillation are used to quantify each control’s contribution to susceptibility ($S_j$, where $j$ corresponds to single layer network). By superposing $S_j$, we create an additive, superposable neural network (SNN) model for total landslide susceptibility. The details of each methodology are explained in the Methods. }
\end{figure}

\newpage
\begin{figure}
\includegraphics[width=0.6\textwidth]{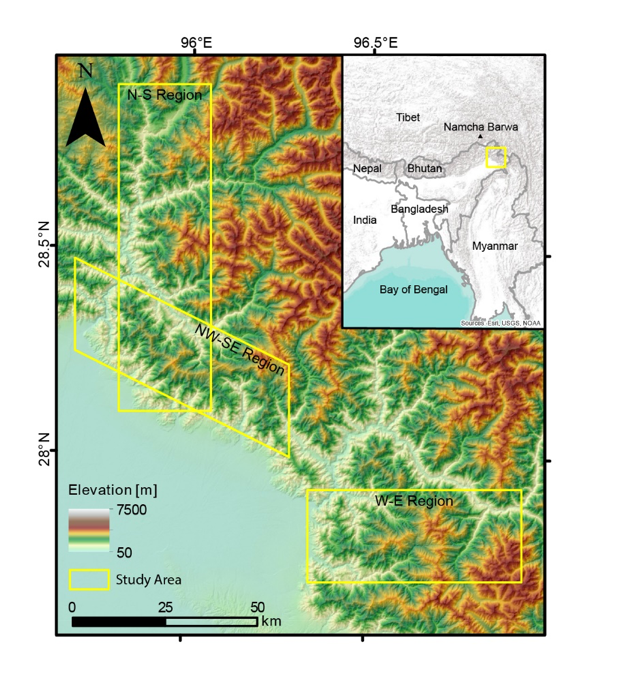}
\caption{\label{fig:studyreg} {\bf Study area in the easternmost Himalaya. } Colors represents the elevation~\cite{EarthExplorer}, and yellow boxes indicate our N-S (Dibang), NW-SE (range front), and E-W (Lohit) oriented study regions. The inset map shows the eastern Himalayan region with our study area shown in a yellow box and national borders shown in dark gray lines.} 
\end{figure}

\newpage
\begin{figure}[h]
	\begin{center}
	\includegraphics[width=\textwidth]{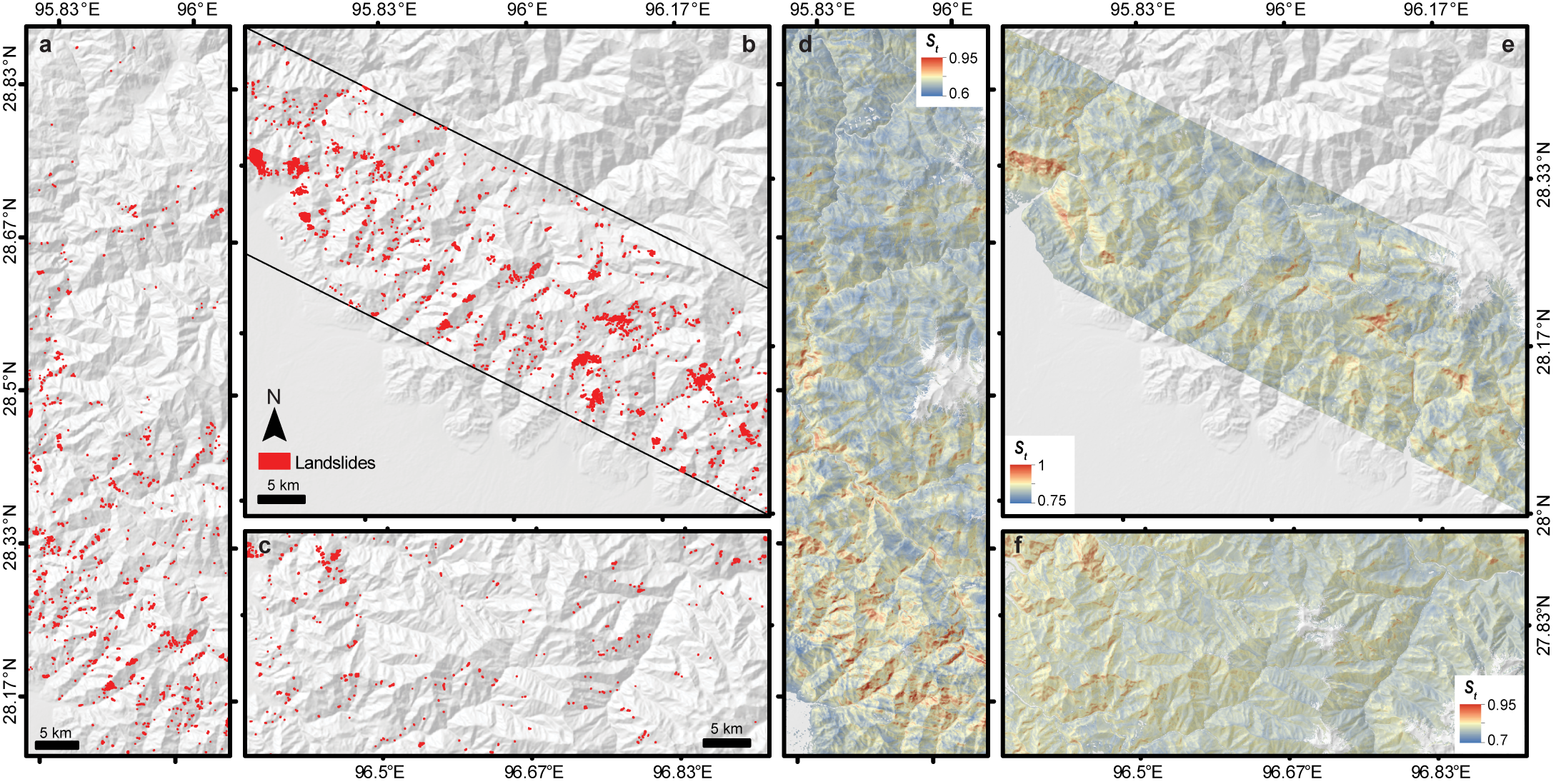} 
	\end{center}
	\caption{\label{fig:R123_LS_060222}  {\bf Mapped landslides and modeled susceptibility.} Spatial distribution of (a-c) mapped landslides and (d-f) modeled landslide susceptibility for the (a,d) N-S, (b,e) NW-SE, and (c,f) E-W study regions. (a) 959, (b) 1536, and (c) 386 landslides are shown in red polygons in (a-c). Total susceptibility at the pixel scale ($S_t$) from the Level-2 superposable neural network are shown in (d-f). The threshold $S_t$ values that are used to classify landslide and non-landslide pixels in the model are (d) 0.767, (e) 0.861, and (f) 0.816, respectively.} 
\end{figure}

\clearpage
\newpage
\begin{figure}[h] %
\includegraphics[width=0.8\textwidth]{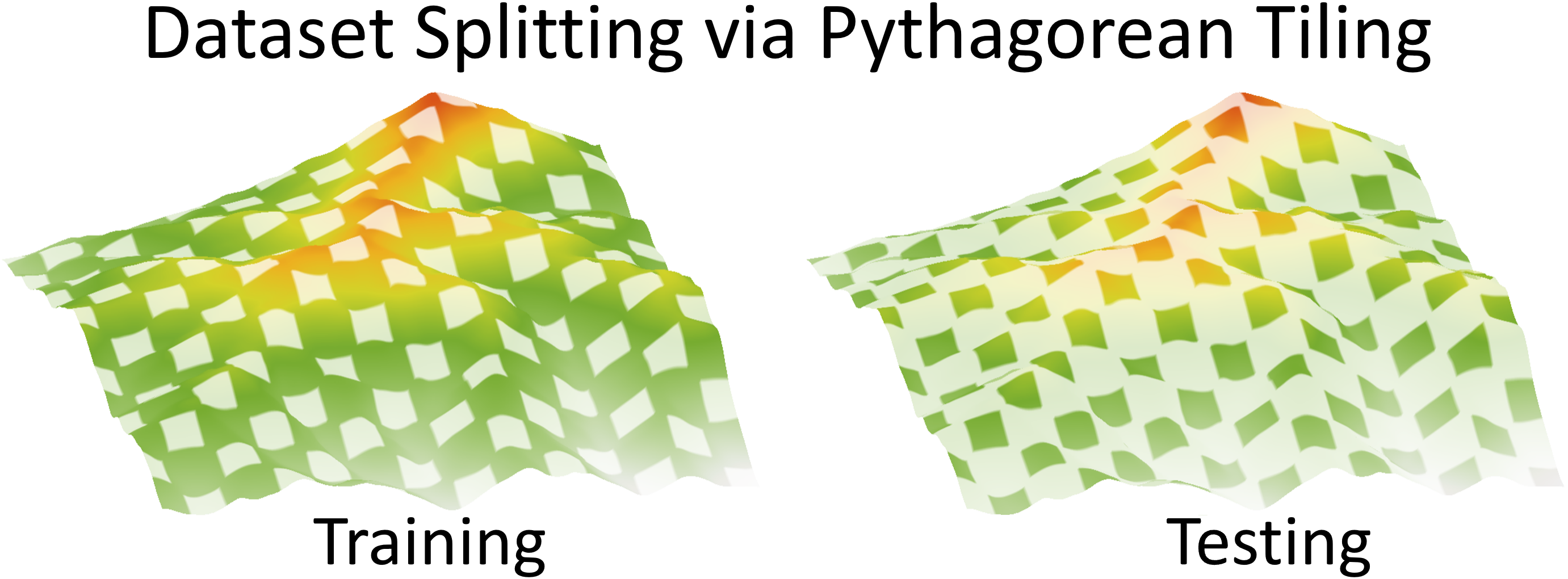}
\caption{\label{fig:PT2}   {\bf Illustration of spatial data partitioning using Pythagorean tiling.} Pythagorean tiling is used to divide data from the modeled region in a spatially representative manner that maintains variability between training and testing partitions. Using Pythagorean tiling, we generate a checkerboard-like pattern with a ~70/30\% square ratio, where bigger squares correspond to training and smaller squares correspond to testing.}
\end{figure}

\newpage
\begin{figure}[h] 
\includegraphics[width=1.0\textwidth]{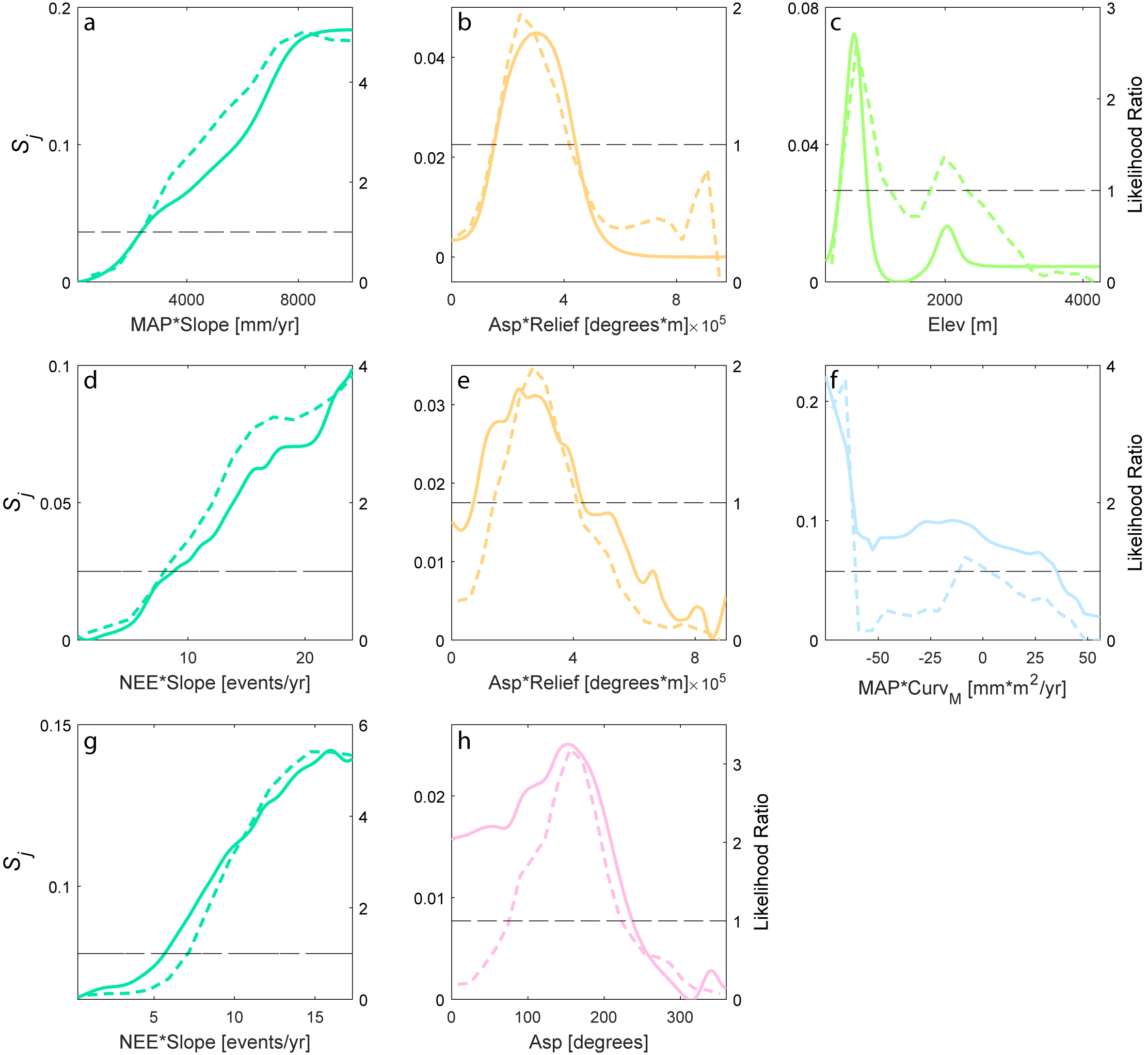}
\caption{\label{fig:All_fvS_052422}  {\bf Individual feature contributions to total susceptibility.} Independent functions of $S_j$ identified as primary landslide controls are shown for the (a-c) N-S, (d-f) NW-SE, and (g, h) E-W study regions. Likelihood ratios ($LR$), representing the likelihood of landslide occurrence for a specific range of feature values, are shown as short, dashed, colored lines with corresponding right-side $y$-axes for reference. $LR$ = 1 and $LR$ $>$ 1 represent the average and above-average likelihood of landslide occurrence, respectively. Note that $S_j$ corresponding to $LR=1$, shown as long-dashed black lines, differ between features because the SNN quantifies the absolute contributions of $S_j$ decoupled from other features. Features related to topography, aspect, climate, and geology are shown in green, pink, blue, and brown or combinations thereof, respectively. Mean annual precipitation (\emph{MAP}), number of extreme rainfall events (\emph{NEE}), aspect (\emph{Asp}), elevation (\emph{Elev)}, mean curvature (\emph{Curv$_M$}), and local relief (\emph{Relief}). The asterisk * indicates algebraic multiplication of two features.} 
\end{figure}

\newpage
\begin{figure}[h] 
\begin{center}
\includegraphics[width=0.8\textwidth]{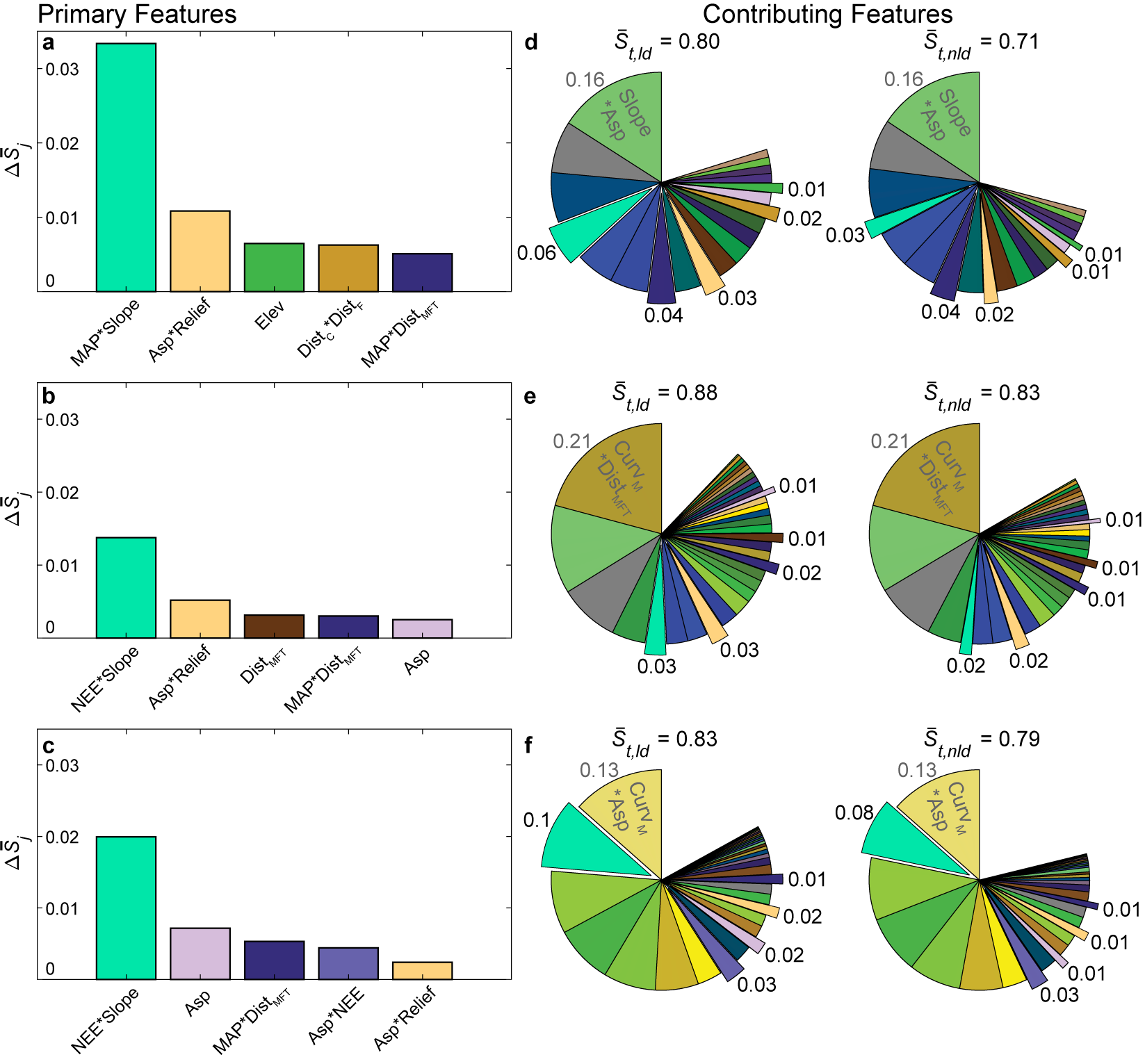} 
\end{center}
\caption{\label{fig:Global_PieBar_052322}  {\bf Feature contributions to total susceptibility} for the (a,d) N-S, (b,e) NW-SE, and (c,f) E-W study regions.  Bar charts in (a-c) represent $\Delta \bar{S}_j$ in descending order, and pie charts in (d-f) represent average $S_j$ ($\bar{S}_j$) contributions to landslide ($ld$) and non-landslide ($nld$) areas. $\Delta \bar{S}_j$ represents the difference in average contribution between areas of $ld$ and $nld$ in each region. Extruding pie chart features are features with large $\Delta \bar{S}_j$) found in the corresponding bar chart on the left. Features related to topography, aspect, climate, and geology are shown in green, pink, blue, and brown or combinations thereof, respectively.  Mean annual precipitation (\emph{MAP}), number of extreme rainfall events (\emph{NEE}), aspect (\emph{Asp}), elevation (\emph{Elev)}, mean curvature (\emph{Curv$_M$}), distances to channel (\emph{Dist$_C$}), all faults (\emph{Dist$_F$}), and the Main Frontal Thrust and suture zone (\emph{Dist$_{MFT}$}), and local relief (\emph{Relief}). The asterisk * indicates algebraic multiplication of two features. Information regarding features is provided in Methods.} 
\end{figure}

\newpage
\begin{figure}[h]%
	\begin{center}
	\includegraphics[width=\textwidth]{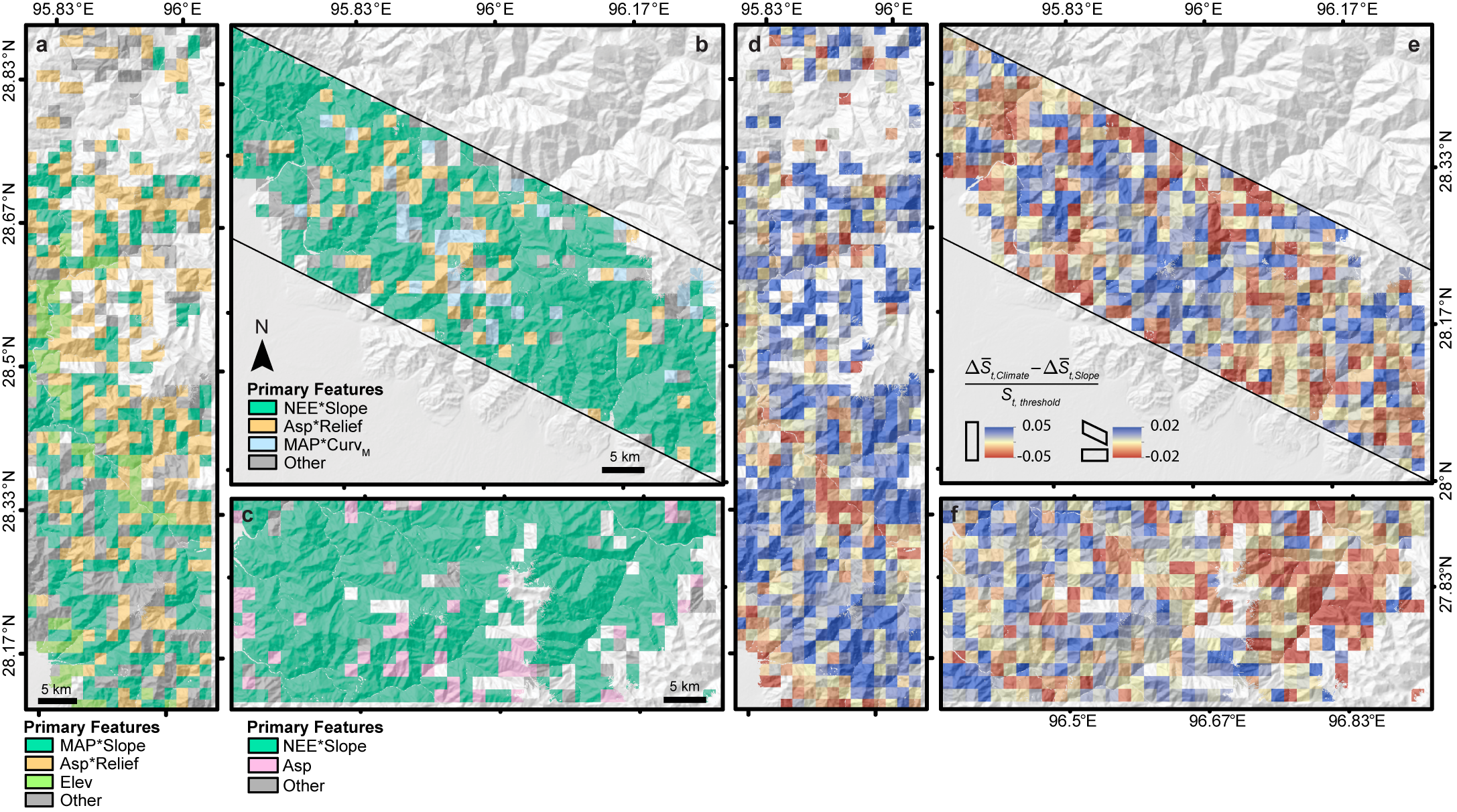} 
	\end{center}
	\caption{\label{fig:R123_Rank_060222}  {\bf  Important controls for landslides.} Spatial distribution of (a-c) primary features identified as locally important controls of landslides and (d-f) relative climate vs slope susceptibility contributions for the (a,d) N-S, (b,e) NW-SE, and (c,f) E-W study regions. The locally important control in (a-c) is identified as the feature with the largest difference in average contribution ($\Delta \bar{S}_j$) between areas of landslides ($ld$) and non-landslides ($nld$) within a 2.25 km\textsuperscript{2} window. The contribution from climate features ($\Delta \bar{S} \textsubscript{\emph{t,Climate}}$, $j$ = \emph{Asp, NEE, MAP}) relative to that of slope ($\Delta \bar{S}\textsubscript{\emph{t,Slope}}$) normalized by the corresponding threshold $S_t$ is shown in (d-f). Windows with a higher climate contribution are colored blue while those with a greater slope contribution are colored red. Windows of no data contain a majority of unmapped areas or indicate lack of modeled landslides. Features related to topography, aspect, climate, and geology are shown in green, pink, blue, and brown or combinations thereof, respectively. Mean annual precipitation (\emph{MAP}), number of extreme rainfall events (\emph{NEE}), aspect (\emph{Asp}), elevation (\emph{Elev)}, mean curvature (\emph{Curv$_M$}), and local relief (\emph{Relief}). The asterisk * indicates algebraic multiplication of two features.}
\end{figure}

\clearpage

\FloatBarrier

\newpage

\begin{centering}
{\bf Supplementary Information for ``Landslide Susceptibility Modeling by Interpretable Neural Network''}
\end{centering}

\noindent\textbf{Contents}
\begin{enumerate}
\item Supplementary Note 1. SNN Validation by Toy Applications
\item Supplementary Note 2. Construction and Performance Assessments of Models
\item Supplementary Note 3. Explanation of Aspect as a Microclimate Control

\item Supplementary Figure 1.  Comparison among the feature ranges of our three study regions
\item Supplementary Figure 2. Flowchart detailing the semi-automatic landslide mapping procedure
\item Supplementary Figure 3. Examples of semi-automatically detected landslides
\item Supplementary Figure 4. Landslide area versus probability density
\item Supplementary Figure 5. Spatial distribution of 15 features used in the superposable neural network model
\item Supplementary Figure 6. The relationship among aspect, normalized difference vegetation index, and ${S}_{Asp}$
\item Supplementary Figure 7. Bar charts representing $\Delta \bar{Sn}_j$ from different methods
\item Supplementary Figure 8. Feature ranking for toy application 1
\item Supplementary Figure 9. Individual feature functions and target and SNN output for toy application 1
\item Supplementary Figure 10. Target and SNN output with noise for toy application 1
\item Supplementary Figure 11. Feature ranking for toy application 2
\item Supplementary Figure 12. Individual feature target functions and SNN results for toy application 2

\item Supplementary Table 1. Description of Landslide Inventory
\item Supplementary Table 2. Description and Ranges of 15 Single Features
\item Supplementary Table 3. AUROC of Models and Single Features
\item Supplementary Table 4. Artificial Neural Network and Statistical Model Confidence Intervals
\item Supplementary Table 5. Performance Metrics for Models
\item Supplementary Table 6. Truth table 
\item Supplementary Table 7. Composite features 
\item Supplementary Table 8. Correlation Metrics Between Features (R-value)
\item Supplementary Table 9. Logistic Regression Control Coefficients

\end{enumerate}

\newpage

\setcounter{section}{0}
\renewcommand{\thesection}{Supplementary Note \arabic{section}}

\section{\bf SNN Validation by Toy Applications} 

There can be many solutions of models that can fit a dataset generated by another model with varying degrees of accuracy. In order to validate our SNN approach, we test it on toy applications with a known solution.

\subsection{\bf Toy Application 1} 
Consider the following constrained toy application by generating a dataset that represents a logical relationship and testing the behavior of our algorithm: 

Take the equation 
\begin{equation}
y= x_1*x_2+x_3*x_4-2*x_1*x_2*x_3*x_4,
\label{eq:y}
\end{equation}
where $x_1, x_2, x_3$, and $x_4$ are Boolean values. It is easy to check that the equation represents the logical relationship 
$$y=(x_1 \wedge x_2)  \vee (x_3 \wedge x_4) \wedge (\overline{x_1 \wedge x_2 \wedge x_3 \wedge x_4})$$
where the truth table is shown in Supplementary Table~\ref{tab:1}.

We generated 1,000 random realizations of $x_1(n)$, $x_2(n)$, $x_3(n)$ and $x_4(n)$ and we calculate the corresponding value $y(n)$ for each of these realizations, where $n=1:1000$.  We tested our algorithm by training an SNN using $x_1(n)$, $x_2(n)$, $x_3(n)$, and $x_4(n)$ as the input and $y(n)$ as the target output, so as to test whether our method can infer the logical relationship from the basic components using only the generated data samples. Up to Level-4 composite features were used in this analysis, for a total number of 15 features as shown in Supplementary Table~\ref{tab:2}.

The resulting feature ranking (Supplementary Figure~\ref{fig:k1}) shows that our algorithm was able to successfully isolate the composite features that exist in the relationship. The truth table (Supplementary Table~\ref{tab:1}) reveals that the higher ranking given to $\{x_1*x_2*x_3*x_4\}$ corresponds to the fact that this feature can decisively explain 50\% of the logical relationship independently from the other features. If $\{x_1*x_2*x_3*x_4\}$ is one, then $y$ is always zero. On the other hand, the other features cannot decisively determine any part of the outcome on their own, but they can decisively determine the outcome if they depend on $\{x_1*x_2*x_3*x_4\}$.

Our method was able to find a solution to the logical relationship that accurately matches the target output (Supplementary Figure~\ref{fig:k2}). Our model is given by
$$ \tilde{y}=f(x_1*x_2)+f(x_3*x_4)+f(x_1*x_2*x_3*x_4), $$
where 
$$ f(x_1*x_2) \approx \begin{cases}
0.1, & \mbox{for~} x_1*x_2=0 \\
1.1, & \mbox{for~} x_1*x_2=1
\end{cases} $$

$$ f(x3*x4) \approx \begin{cases}
-0.25, & \mbox{for~} x_3*x_4=0 \\
0.75, & \mbox{for~} x_3*x_4=1
\end{cases} $$

$$ f(x_1*x_2*x_3*x_4) \approx \begin{cases}
0.15, & \mbox{for~} x_1*x_2*x_3*x_4=0 \\
-1.85, & \mbox{for~} x_1*x_2*x_3*x_4=1
\end{cases} $$
Note that subtracting the two ends of the function of each feature returns the coefficient values in the original equation (Eq.~\ref{eq:y}) for each corresponding variable:
$$ f(x_1*x_2) : 1.1-0.1=1, $$
$$ f(x_3*x_4) : 0.75 - (-0.25)=1, $$
$$ f(x_1*x_2*x_3*x_4) : -1.85-0.15= -2. $$

Furthermore, the behavior of our method proved to be robust to noise. Supplementary Figure~\ref{fig:k3} demonstrates the results of the same experiment, but here the data was deliberately contaminated by adding Gaussian noise to $x_1(n)$, $x_2(n)$, $x_3(n)$, and $x_4(n)$ prior to training and testing. 
Although the SNN output became noisier, the noise did not affect the overall outcome and could easily be removed by thresholding 

\subsection{\bf Toy Application 2} 
In the next toy application, consider three features $x_1, x_2$, and $x_3$. 
We generate three functions $f(x_1), f(x_3)$, and $f(x_1*x_3)$ 
and take their sum $y = f(x_1) + f(x_3) + f(x_1*x_3)$. By training an SNN to estimate y using $x_1$, $x_2$, and $x_3$ as inputs, we test whether it can retrieve the contributing composite features and their functions as an interpretation of its solution. The SNN is trained using 7000 randomly generated examples and is tested using another 3000 randomly generated samples. The results shown in Supplementary Figure~\ref{fig:k4} and Supplementary Figure~\ref{fig:k5} demonstrate the ability of the SNN model to perfectly retrieve the individual contributing features and their functions.

\section{\bf Construction and Performance Assessments of Models} 
We evaluated the performance of the SNN compared to traditional approaches using several performance metrics including the area under the receiver operating characteristic curve (AUROC), 
accuracy, sensitivity (i.e., probability of detection, POD), specificity (i.e., probability of false detection, POFD), and POD-POFD following the literature.  See for example ~\cite{prakash2020mapping}. AUROC 
is a cutoff-independent performance criteria while accuracy, POD, and POFD are cutoff-dependent. The AUROC is calculated as the area under a curve created by plotting the true positive rate against the false positive rate at various thresholds along a feature's range. 
AUROC ranges between 0 and 1, with 1 indicating a perfect classifier and 0.5 indicating a random model. After generating a threshold-modeled landslide map based within the $\sim$30\% testing partition using the optimal $S_t$ threshold corresponding to the point closest to [0,1] on an ROC curve, accuracy is calculated as the fraction of landslide and non-landslide area correctly classified by the model relative to all studied areas. POD and POFD measure the proportion of landslide areas correctly classified relative to all observed landslide areas and the proportion of incorrectly classified landslide areas relative to all observed non-landslide areas, respectively.

We calculated these metrics for all 15 single features, a physically-based slope stability model (SHALSTAB), two statistical methods (logistic regression and likelihood ratios), 
and Level-1 and Level-2 SNNs. First, we investigated each of the 15 single features as individual classifiers for landslide occurrences (Supplementary Table~\ref{tab:S3}). Second, we assessed the propensity of landslides using a topographic metric called the failure index. The failure index (\emph{FI}) is the ratio of driving to resisting forces on a hillslope, which is the inverse of the factor-of-safety. \emph{FI} is modified from SHALSTAB, which couples infinite slope stability and steady-state hydrology for a cohesionless material~\cite{Montgomery1994,Dietrich1995,Dietrich2001,Moon2011}. 
Considering that landslides smaller than 100,000~m$^2$ (the upper bound for soil landslides found from global and Himalayan landslide compilations~\cite{Larsen2010, Larsen2012}) constitute $>$99\% of landslides in number and {\raise.17ex\hbox{$\scriptstyle\sim$}}80\% of total landslide area, we assumed that most landslides in our inventory are soil landslides. 

To calculate the \emph{FI}, we first determined the spatial distribution of wetness ($W$), which represents the degree of subsurface saturation. $W$ is calculated as the ratio between local hydraulic flux from a given steady-state precipitation relative to that of soil profile saturation~\cite{Montgomery1994}:
\begin{align}\label{eq:5} 
    W = \frac{h}{z} = \frac{qA}{bT\sin\theta}
\end{align}
where $h$ is the saturated height of the soil column ($L$), $z$ is the total height of the soil column ($L$), $q$ is the steady-state precipitation during a storm event ($L$/$T$), $A$ is the drainage area (\emph{L\textsuperscript{2}}) draining across the contour length $b$ ($L$), $T$ is the soil transmissivity when saturated ($L\textsuperscript{2}/T$), and $\theta$ is the local slope in degrees.  $W$ varies from 0 (unsaturated) to a capped value of 1 (fully saturated). We used the spatial distribution of \emph{MAP}~\cite{BookhagenBurbank2010} to represent the steady-state precipitation, $q$. The $T$ value may vary spatially depending on surface conditions such as depth of soil or weathered rock and hydraulic conductivity ~\cite{montgomery2002piezometric}. However, we do not have field measurements to constrain the spatial variation of this value. Very high or low $T$ values will result in spatially uniform wetness values of 0 or 1, respectively. Thus, we used a base value of 1$\times$10$^{-4}$~m\textsuperscript{2}/s for $T$ following Moon et al.~\cite{Moon2011}, which allowed for a large spatial variation of wetness influenced by precipitation gradient across the area. We then calculated the spatial distribution of \emph{FI} as:
\begin{align}\label{eq:6} 
    FI = \frac{S}{S_0} \left( 1 - W \frac{\rho_w}{\rho_s} \right)^{-1} 
\end{align}
where $S_0$ is the threshold slope set at 45$^{\circ}$, $S$ is the local slope, $\rho_s$ is the wet bulk density of soil (2.0 g/cm$^3$), and $\rho_w$ is the bulk density of water (1.0 g/cm$^3$). To examine whether the performance of \emph{FI} is different when predicting all landslides vs soil landslides, we included the performance metric results for \emph{FI} calculated using all landslides and soil landslides in Supplementary Table~\ref{tab:S3}.

Third, we applied two statistical models, logistic regression and likelihood ratios, to assess landslide susceptibility. Logistic regression (hereafter, LogR) is based on a multivariate regression between a binary response of landslide occurrence and a set of predicting features that are continuous, discrete, or a combination of both types~\cite{Lee2005}.  To build these models, we considered only one curvature metric following Lee~\cite{Lee2005}, instead of using all four different curvatures. We selected \emph{Curv$_M$} to build the statistical models. In addition, we considered log$_{10}$(drainage area) and log$_{10}$(discharge) because of their inverse power-law relationships with landslide and debris flow incision~\cite{stock2003valley, stock2006erosion}.   The relationship between features and landslide occurrence can be displayed as:
\begin{align}\label{eq:7} 
    p = \frac{e^c}{e^c+1}
\end{align}
where \emph{p} is the probability of landslide occurrence that varies from 0 to 1 in an S-shaped curve, and \emph{c} is the linear combination of features:
\begin{align}\label{eq:8} 
    c = b_0 + b_1x_1 + b_2x_2+...+b_nx_n
\end{align}
where $x_i$ ($i$ = 1, 2, . . ., $n$) represents each feature, $b_i$ represents the optimized coefficient, and $b_0$  represents the intercept of the model. Utilizing Eqs.~\ref{eq:7} and \ref{eq:8}, we obtained an extended expression for the LogR model relating the probability of landslide occurrence $p$ and multiple features:
\begin{align}\label{eq:9} 
    \mbox{logit}(p) = \log \left(\frac{p}{1-p} \right) = b_0 + b_1x_1 + b_2x_2+...+b_nx_n
\end{align}
where $\log$ is the natural log. To determine any possible collinearity between features, we calculated the correlation coefficient ($R$) between all combinations of 12 features (Supplementary Table~\ref{tab:S6}). We observed maximum absolute values of $R$ = -0.828 (N-S), 0.717 (NW-SE), and 0.857 (E-W), which are below the threshold of 0.894 corresponding to a variance inflation factor of \textless{5}. $R$ below this threshold indicate low collinearity between features~\cite{Stine1995,Kavzoglu2014} and thus we used all 12 features. We treated aspect as a discrete feature due to its nonlinear relation with landslide occurrences. The best-fit coefficient values are shown in Supplementary Table~\ref{tab:S7}.

Similar to the SNN, the LogR method provides information about the importance of variables through the best-fit coefficients. To compare those results, we determined top features that differentiate areas with and without threshold modeled landslides for the N-S region based on the LogR output following similar procedures that we used for the SNN. The output of LogR ranges between 1.06$\times$10$^{-6}$ to 0.820. The threshold value ($t$) of 0.005 that corresponds to the point closest to [0,1] on an ROC curve (i.e., a perfect classifier) was used to classify landslide (\emph{ld}) and non-landslide areas (\emph{nld}) for the N-S region. We calculated $\Delta \bar{cn}_{i}$ as the difference between the average value of a feature multiplied by its respective coefficient for \emph{ld} and \emph{nld} areas, then divided by an adjusted threshold that was transformed from $t$ (i.e., 0.005) from the LogR output according to the equation below:
\begin{align}\label{eq:8b} 
    \Delta \bar{cn}_{i} = \frac{\bar{c}_{i,ld} - \bar{c}_{i,nld}}{t_a}\,\mbox{~where~}\, t_a = \log \left( \frac{t}{1-t} \right)-b_0,
\end{align}
where $\bar{c}_{i,a}$ is the average feature ($i$) value multiplied by its respective coefficient for areas ($a$) of \emph{ld} or \emph{nld}, and $t_a$ is the adjusted threshold value based on $b_0$, the overall intercept value determined by the LogR model and $t$, the threshold determined using the ROC curve (i.e., 0.005 for the N-S region). We transformed $t$ to $t_a$ and used it for normalization to enable the direct comparison of results between LogR and the SNN. For the SNN-determined primary features, we calculated $\Delta \bar{Sn}_j$ as $\Delta \bar{S}_j$ divided by the threshold that is used to classify landslides (i.e., 0.767 for SNN Level-2 and 0.399 for SNN Level-1 for the N-S region).  \label{sec:dsnj}
The value of 1 in both $\bar{Sn}$ and $ \bar{cn}$ represents the threshold susceptibility that classifies \emph{ld} and \emph{nld} areas. The results of the identified primary controls of landslides, which induce large differences in average susceptibility between \emph{ld} and \emph{nld} areas, in the N-S region from LogR, the SNN Level-1, and the SNN Level-2 are shown in Supplementary Figure \ref{fig:M_Sweight}. All methods identified climate-related factors (e.g., \emph{MAP}, \emph{NEE}, \emph{Asp}) as primary controls; however, only the SNN Level-2 was able to identify the importance of the composite feature \emph{MAP}*\emph{Slope}.

The likelihood ratio method uses the relationship between observed landslide occurrences and controlling feature ranges. Previous studies have quantified the ratio of the probability of landslide occurrences within a range of feature values to the probability of non-occurrences or all-occurrences and referred to it as the likelihood ratio, frequency ratio, or probability ratio~\cite{Lee2005,akgun2012comparison,reichenbach2018review}.  In this study, we calculated likelihood ratios ($LR$) as the ratio of the percentage of landslide pixels relative to total landslide pixels divided by the percentage of pixels relative to the total area for a specific range of feature values~\cite{Lee2005,akgun2012comparison}. 
Landslide susceptibility for each pixel is calculated as the sum of the corresponding $LR$ from each feature’s value. A ratio of 1 and $>$1 indicates the average and above-average likelihood of landslide occurrence within the feature range compared to that of the study area. Conversely, values less than 1 indicate a below-average likelihood. In this study, we used all 15 single features with each feature's range divided into ten bins to calculate $LR$ and landslide susceptibility. The first and last bins represent areas less than and greater than the 10\textsuperscript{th} and 90\textsuperscript{th} percentile of $LR$, respectively, with values between these bins split into eight equal bin ranges.

We determined 95\% confidence intervals of mean AUROC by conducting a 10-fold cross validation for all statistical and neural network models utilized in this study. 
We tested the trained model on 50\% of the testing dataset that was selected randomly and uniformly. We then calculated the AUROC for each trial. This procedure was repeated 10 times for each method and the results were used to calculate the 95\% confidence interval for the mean AUROC and $\pm$ 2$\sigma$ range of AUROC from 10 validation tests (Supplementary Table~\ref{tab:S4}). 

Our model assessments for the single features indicate that \emph{MAP} [AUROC = 0.756 (N-S region)] and slope [AUROC = 0.696 (NW-SE), 0.760 (E-W)] are the highest performing single features. The SNN produces a $\sim$19-22\% average improvement in AUROC compared to a physically-based landslide model (e.g., failure index for all landslides or soil landslides). The physically-based model of \emph{FI} produces slightly different AUROC when predicting all landslides vs. soil landslides, but both AUROC values were lower than that of the SNN (Supplementary Table~\ref{tab:S3}). 
Additionally, the SNN produced an average of $\sim$5\% and $\sim$8\%
increases in performance compared to the LogR and $LR$ methods, 
respectively. Further investigation using performance metrics including the AUROC, 
accuracy, POD, POFD, and POD-POFD reveals that the SNN largely outperformed the tested 
statistical and physically-based models across all metrics (Supplementary Tables ~\ref{tab:S3} and ~\ref{tab:S5}).

Our implementation of \emph{FI} can be improved by including additional model parameters (e.g., cohesion), calibrating parameters such as soil depth or transmissivity to account for landscape heterogeneities, performing parameter optimizations, or adopting probabilistic approaches in future studies (e.g.,~\cite{raia2014improving}). To properly calibrate model parameters, we need extensive field measurements, which are not currently available. Without field-calibrated model parameters, physically-based models often yield lower performance compared to data-driven models at a regional scale (e.g., ~\cite{yilmaz2009gis}).
 

\section{{\bf Explanation of Aspect as a Microclimate Control}}

The SNN identified aspect, the direction of slope face, as another primary feature that influences landslide occurrences. Previous studies considered hillslope aspect preference in terms of: 1) vegetation activity that affects root cohesion~\cite{mcguire2016elucidating}, or 2) the orientation of wind-driven rainfall. To examine vegetation activity across hillslope aspect, we calculated the normalized difference vegetation index (\emph{NDVI}) following the USGS procedure~\cite{Zanter2016}.   We first converted Landsat 8 Level-1 Digital Numbers to top-of-atmosphere (TOA) reflectance. TOA reflectance eliminates the impact of different solar angles and illumination geometries and is calculated as:
\begin{align}\label{eq:3} 
    \rho\lambda = \frac{M_\lambda Q_{cal}+A{\rho}}{\cos(\theta_{SZ})}
\end{align}
where $\rho\lambda$ is the TOA reflectance, $M_\lambda$ is the band-specific multiplicative rescaling factor from the Landsat 8 metadata, \emph{Q}$_{cal}$ are the standard product pixel values, $A\rho$ is the band-specific additive rescaling factor from the metadata, and $\theta_{SZ}$ is the local solar zenith angle.\par
We use the corrected bands 4 and 5 from Landsat 8 to calculate NDVI as:
\begin{align}\label{eq:4} 
    \emph{NDVI} = \frac{ \mbox{band}\,5 - \mbox{band}\,4}{\mbox{band}\,5 + \mbox{band}\,4}
\end{align}
where bands 4 and 5 represent visible and near-infrared light reflected by vegetation, respectively. Healthy vegetation with high photosynthetic capacity absorbs a larger proportion of incident visible light while reflecting a greater portion of near-infrared light compared to sparse or unhealthy vegetation~\cite{Tucker1986}.  Therefore, an \emph{NDVI} value close to 1 suggests a higher density of healthy vegetation and green leaves while a value near 0 might indicate unhealthy or no vegetation. We utilized Landsat 8 satellite imagery from October 2015, November 2017, and February 2018~\cite{EarthExplorer} for our analyses of \emph{NDVI}. These months were selected to characterize \emph{NDVI} values before and after the summer monsoon season, during which a large proportion of landslides are suspected to occur because of intense rainfall. We excluded summer months from our analyses because of the abundant cloud cover present in those images, which masks the visibility of the land surface. 

\emph{NDVI} plotted against aspect in our study areas shows a broad distribution of high values centered around values corresponding to south-facing slopes. However, this \emph{NDVI} distribution is different from the observed peak of \emph{S\textsubscript{Asp}} around 145$^{\circ}$  to 180$^{\circ}$  (Supplementary Figure~\ref{fig:s1}). This result may imply that more landslides on south-facing slopes are likely due to orographic precipitation patterns caused by moisture delivery from the south rather than through the effects of vegetation. In fact, if vegetation root cohesion has a substantial impact on landslide stability, we would expect decreased landslide occurrences in south-facing slopes considering the increased \emph{NDVI.} Previous work has characterized the northward moisture transfer to this study area from the Bay of Bengal during monsoon seasons~\cite{BookhagenBurbank2010,Barros2004,yang2018atmospheric}. Thus, we believe that the SNN-identified primary feature aspect supports the influence of aspect-related differences in microclimate (e.g., moisture availability) on landslide occurrences in this area.

\clearpage

\FloatBarrier

\newpage

\setcounter{figure}{0}
\renewcommand{\figurename}{ {\bf Supplementary Figure} }
\renewcommand{\thefigure}{ {\bf \arabic{figure}} }

\newpage

\begin{figure}
\includegraphics[width=0.5\textwidth]{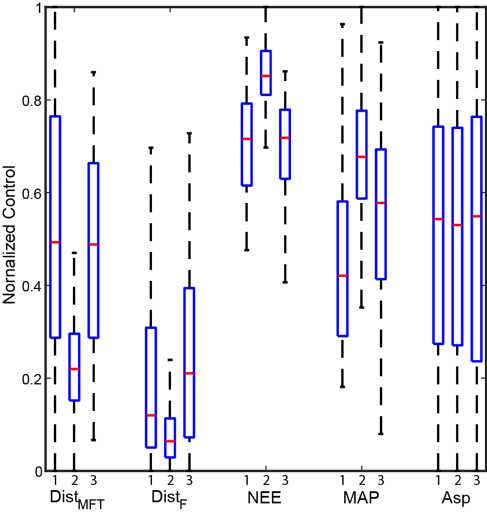}
\caption{\label{fig:M_S4} {\bf Comparison among the feature ranges of our three study regions.} Feature ranges of distance to the Main Frontal Thrust and suture zone (\emph{Dist\textsubscript{MFT}}), distance to all faults (\emph{Dist\textsubscript{F}}), number of extreme rainfall events (\emph{NEE}), mean annual precipitation (\emph{MAP}), and aspect (\emph{Asp}), each normalized by the maximum feature value across all three regions. Red center lines represent the median and top and bottom ends of the box represent the 25\textsuperscript{th} and 75\textsuperscript{th} quartiles, respectively. The ends of the dashed lines extending from each side of the box plot represent 1.5 times the interquartile range or the minimum or maximum values. 
On the $x$-axis, 1, 2, and 3 correspond to the N-S (Dibang), NW-SE (range front), and E-W (Lohit) regions, respectively.}
\end{figure}

\newpage
\begin{figure}
\includegraphics[width=1.0\textwidth]{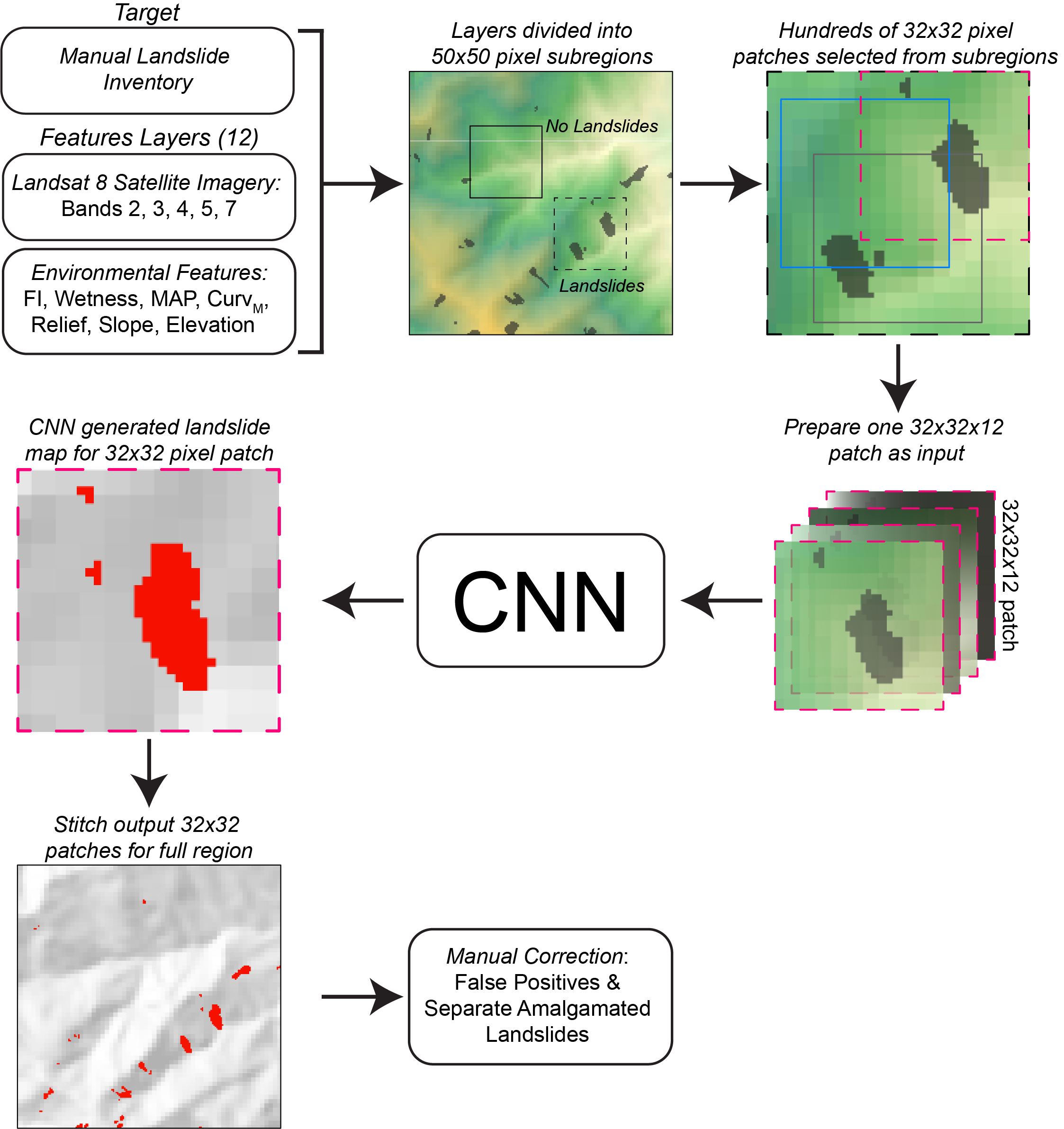}
\caption{\label{fig:CNNflowchart} {\bf Flowchart detailing the semi-automatic landslide mapping procedure}.}
\end{figure}

\newpage
\begin{figure}
\includegraphics[width=0.85\textwidth]{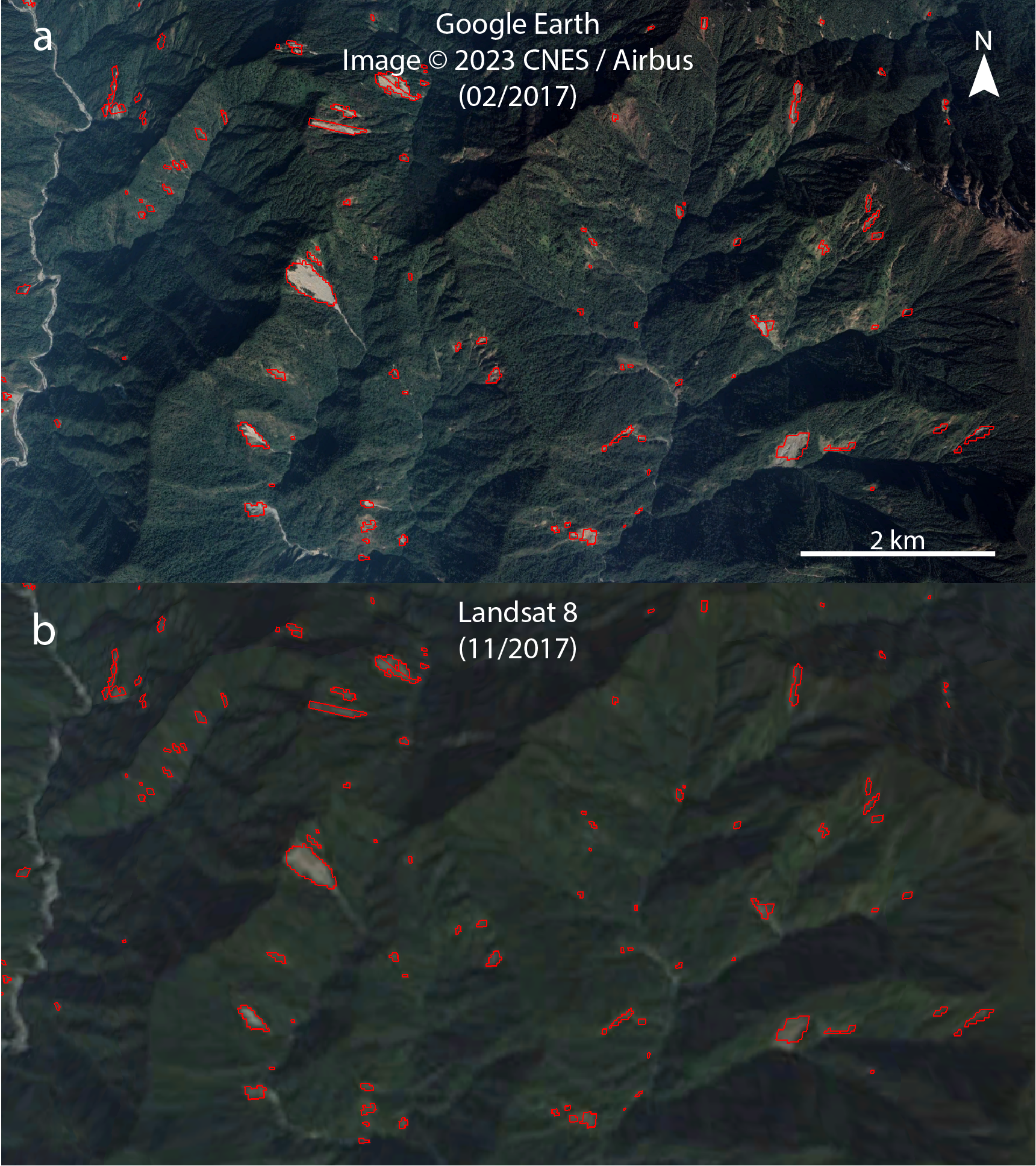}
\caption{\label{fig:GELSchar} {\bf Examples of semi-automatically detected landslides.} Mapped landslide polygons from the N-S (Dibang) subregion are shown in red outlines with background images from (a) Google Earth and (b) Landsat 8. Landsat 8 natural imagery is composed of bands 2, 3, and 4, but landslide mapping is based on 5 bands (2, 3, 4, 5, 7) and 7 input features (see Methods).}
\end{figure}

\newpage

\begin{figure}
\includegraphics[width=0.5\textwidth]{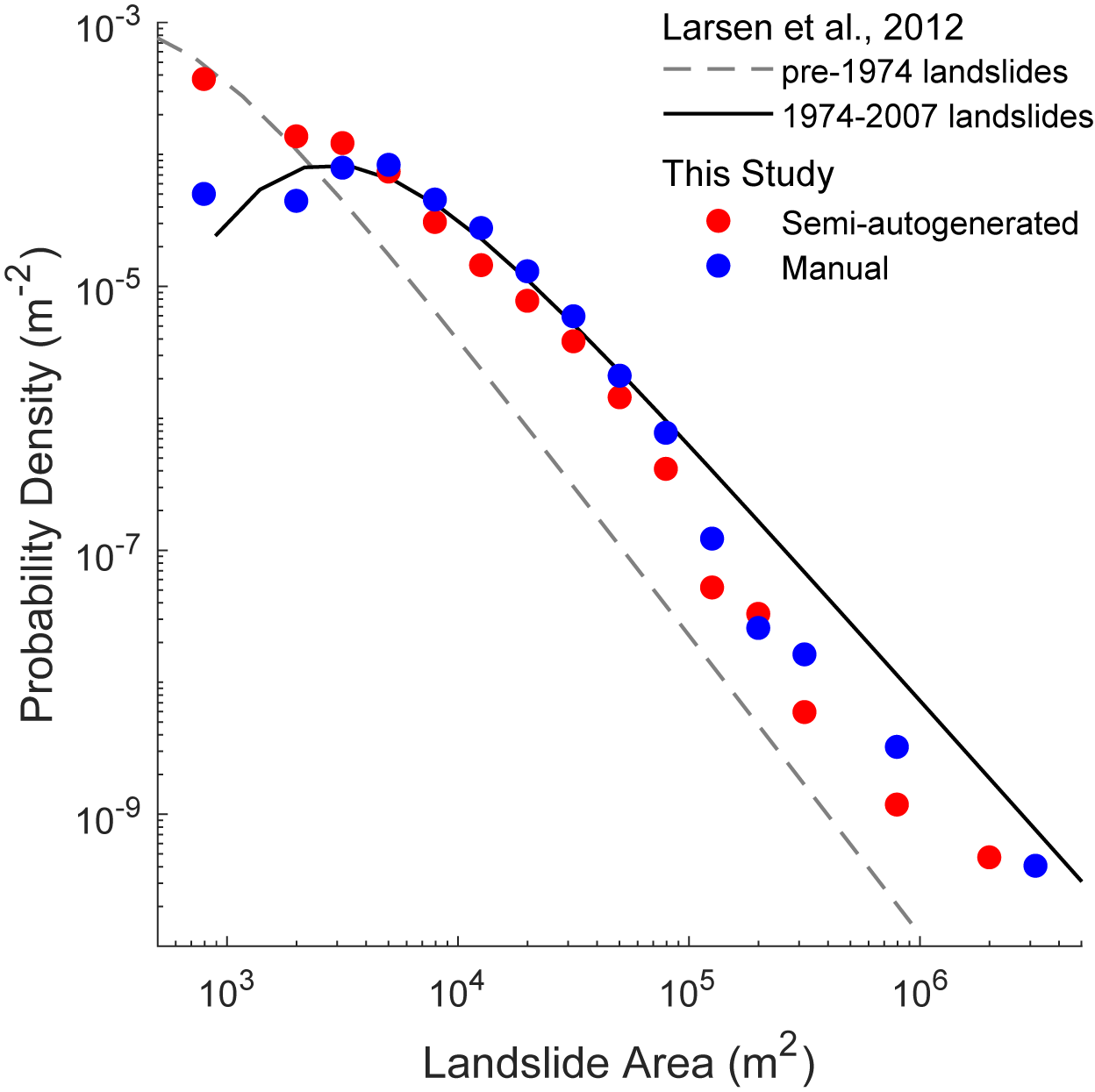}
\caption{\label{fig:M_areafreq} {\bf Landslide area versus probability density.} The manually and semi-automatically mapped landslides before 2017 from our site are shown in blue and red circles, respectively. For reference, the inverse-gamma fits of the pre-1974 (grey dashed line) and 1974-2007 (black solid line) landslides from the nearby Namche Barwa region in the eastern Himalaya~\cite{Larsen2010,Larsen2012} are shown.}
\end{figure}

\newpage

\begin{figure}
\includegraphics[width=0.50\textwidth]{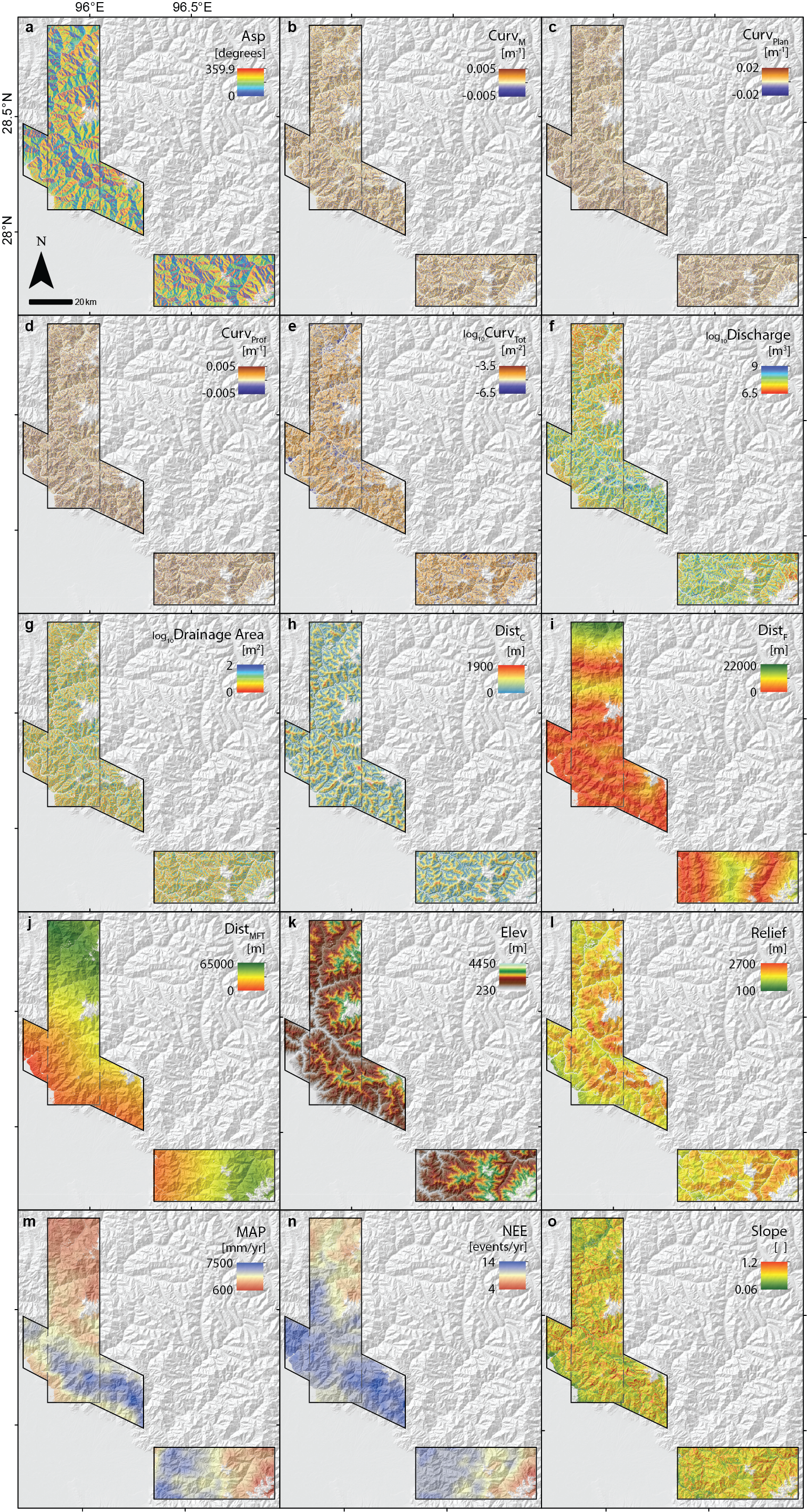} 
\caption{\label{fig:M_S3} {\bf Spatial distribution of 15 features used in the superposable neural network model.} The 15 single features include (a) aspect (\emph{Asp}), (b) mean curvature (\emph{Curv}$_M$), (c) planform curvature (\emph{Curv}$_{Plan}$), (d) profile curvature (\emph{Curv}$_{Prof}$), (e) total curvature (\emph{Curv}$_{Tot}$), (f) discharge, (g) drainage area, (h) distance to channel (\emph{Dist}$_C$), (i) distance to faults (\emph{Dist}$_F$),  (j) distance to the Main Frontal Thrust and suture zone (\emph{Dist}$_{\emph{MFT}}$), (k) elevation (\emph{Elev}), (l) local relief (\emph{Relief}), (m) mean annual precipitation (\emph{MAP}), (n) number of extreme rainfall events (\emph{NEE}), and (o) slope. Dashed lines mark the overlapping area between the N-S (Dibang) and NW-SE (range front) regions. Features in (e, f, g) are displayed on logarithmic scales.}
\end{figure}

\newpage 
\begin{figure}
\includegraphics[width=0.6\textwidth]{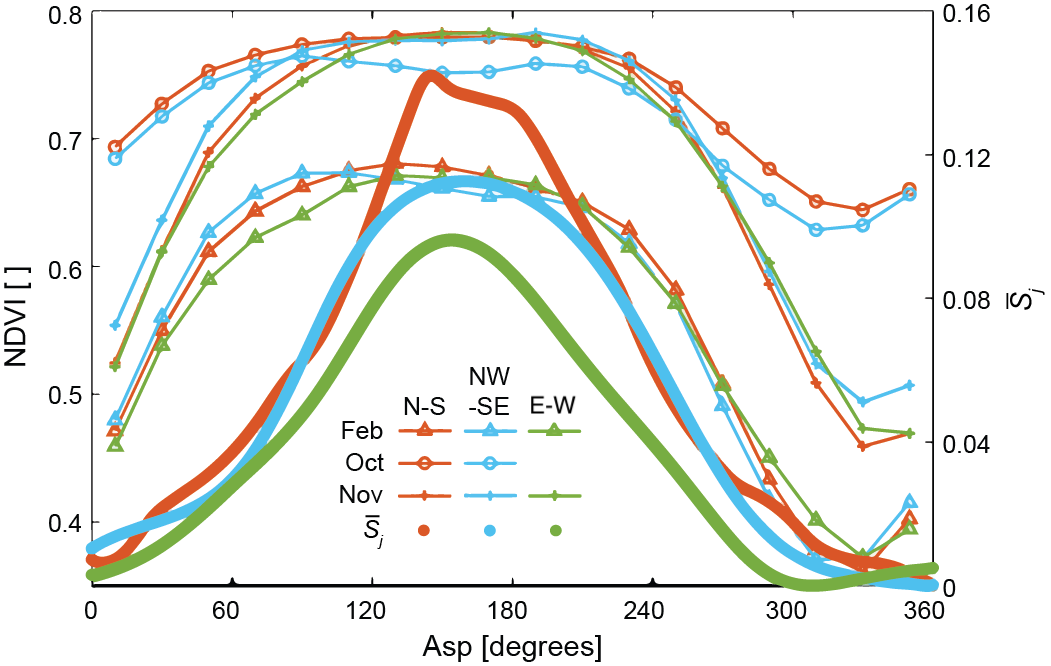}
\caption{\label{fig:s1} {\bf The relationship among aspect, normalized difference vegetation index, and $S\textsubscript{\emph{Asp}}$.} The normalized difference vegetation index (\emph{NDVI}) is shown in  thin lines and $S\textsubscript{\emph{Asp}}$ from SNN Level-1 is shown in thick lines. Colors correspond to different regions while symbols shown as thin lines correspond to different times of measurement (October 2015, November 2017, and February 2018). Symbols on thin lines represent the averaged \emph{NDVI} value for a 20$^{\circ}$ interval of aspect. N-S (Dibang), NW-SE (range front), E-W (Lohit).}
\end{figure}

\newpage
\begin{figure}
\includegraphics[width=1.0\textwidth]{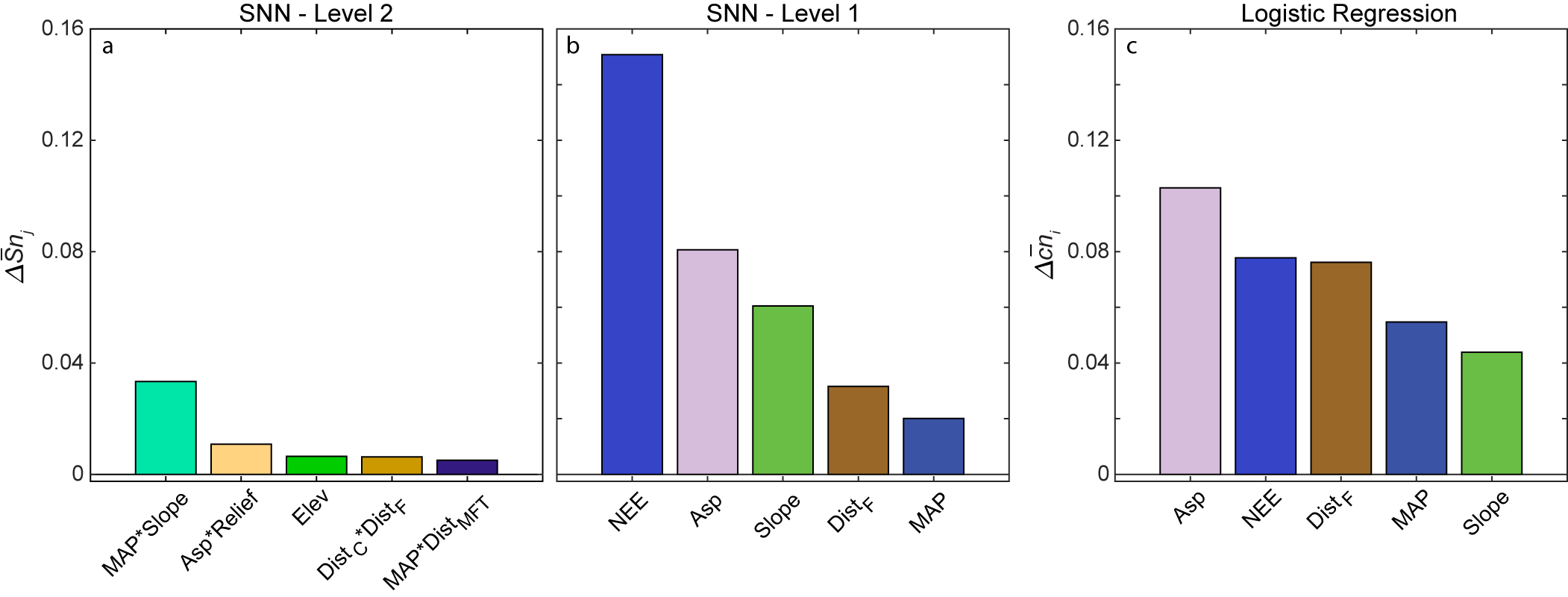}
\caption{\label{fig:M_Sweight} {\bf Bar charts representing $\Delta \bar{Sn}_j$ from different methods.} $\Delta \bar{Sn}_j$ from the (a) SNN Level-2 and (b) SNN Level-1 and $\Delta \bar{cn}_{i}$ for (c) logistic regression for the N-S region, arranged in descending order. Details on the calculations of $\Delta \bar{Sn}_j$ and $\Delta \bar{cn}_{i}$ are provided in Supplementary Note 2, page~\pageref{sec:dsnj}. Features related to topography, aspect, climate, and geology are shown in green, pink, blue, and brown or combinations thereof, respectively. Mean annual precipitation (\emph{MAP}), number of extreme rainfall events (\emph{NEE}), aspect (\emph{Asp}), elevation (\emph{Elev)}, mean curvature (\emph{Curv$_M$}), distances to channel (\emph{Dist$_C$}), all faults (\emph{Dist$_F$}), and the Main Frontal Thrust and suture zone (\emph{Dist$_{MFT}$}), and local relief (\emph{Relief}). The asterisk * indicates algebraic multiplication of two features.}
\end{figure}

\newpage
\begin{figure}
\includegraphics[width=0.75\textwidth]{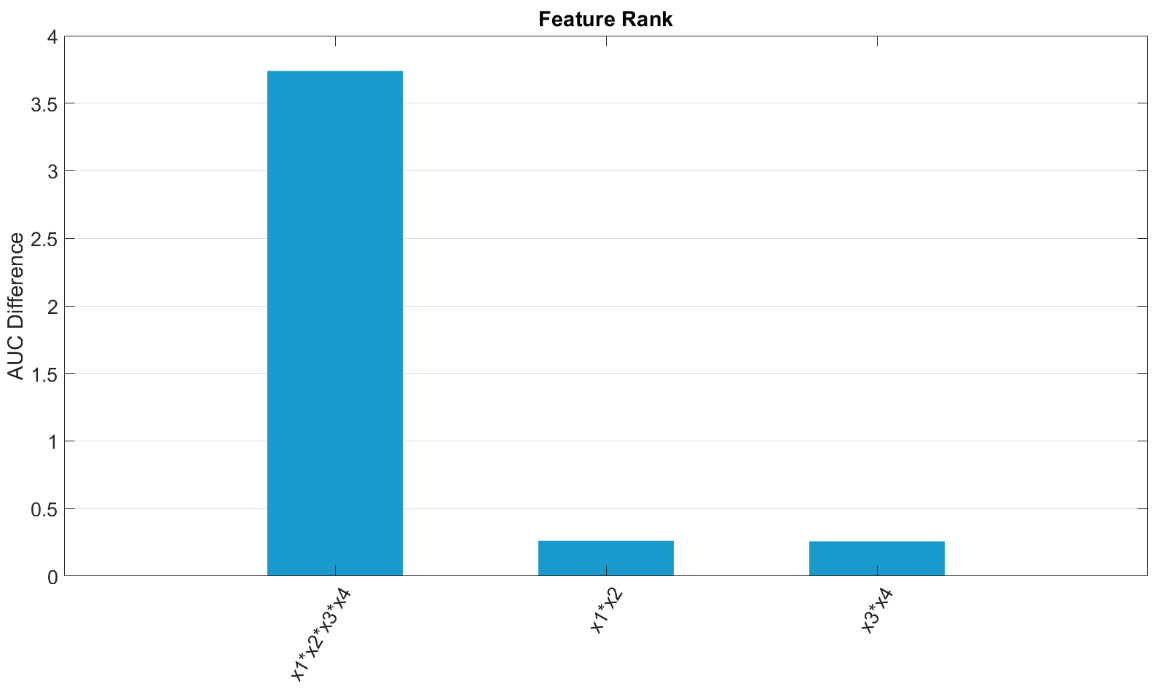}
\caption{\label{fig:k1} {\bf Toy application 1: Feature ranking.}}
\end{figure}

\newpage

\begin{figure}
\includegraphics[width=0.75\textwidth]{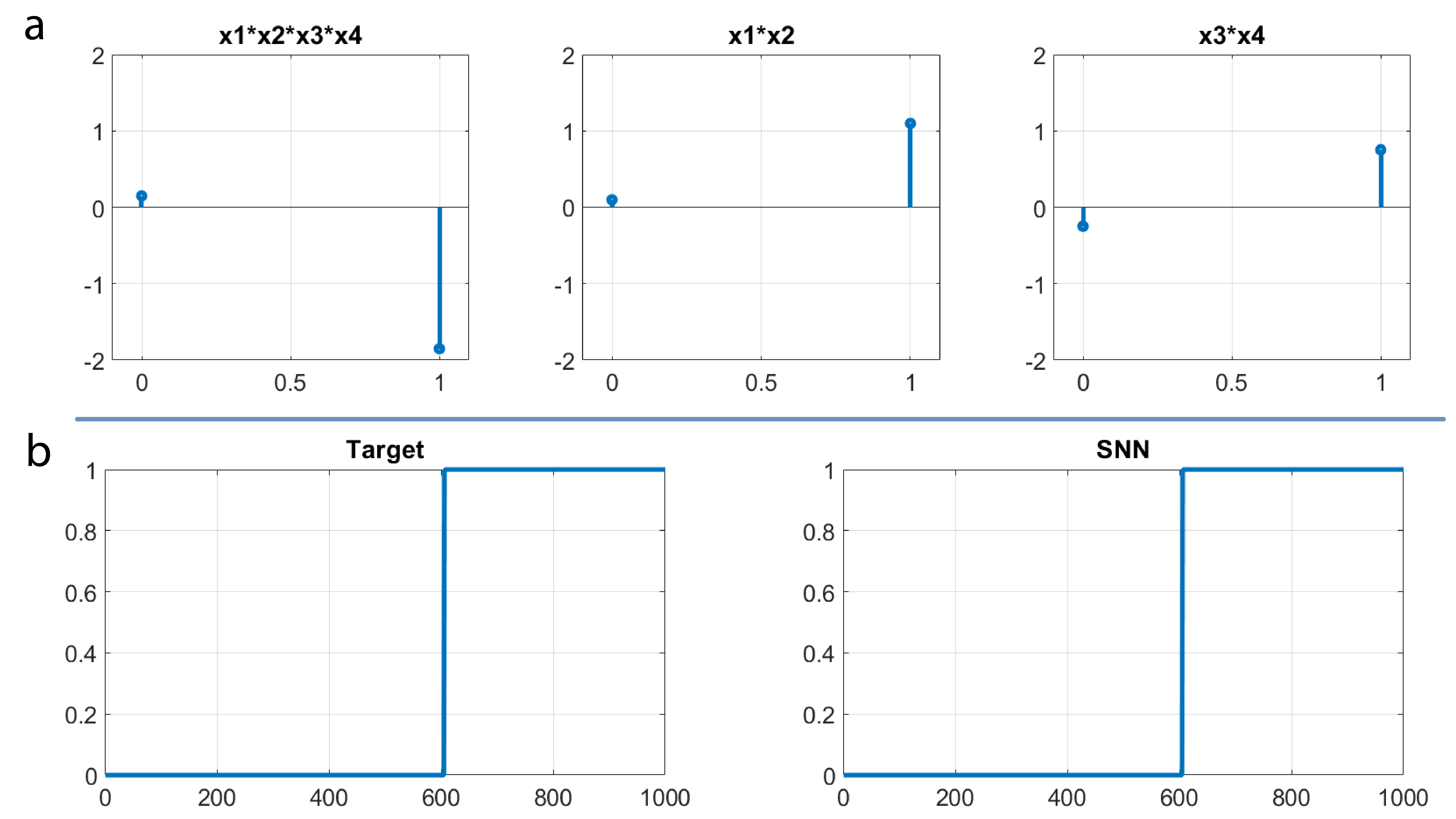}
\caption{\label{fig:k2} Toy application 1: (a) individual feature functions. (b) Target output v.s. SNN output.}
\end{figure}

\newpage

\begin{figure}[h]
\includegraphics[width=0.75\textwidth]{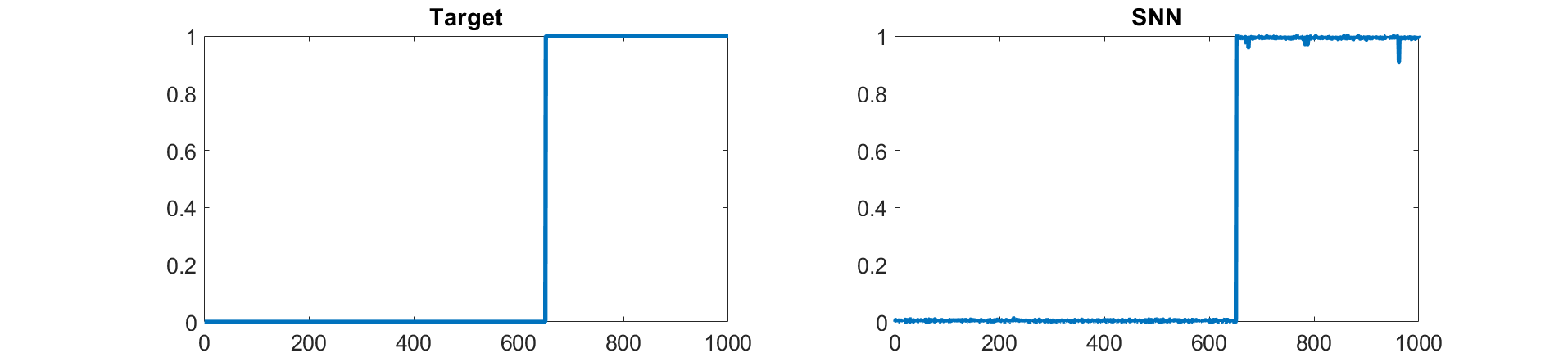}
\caption{\label{fig:k3} Toy application 1: Target output v.s. SNN output when data is contaminated with noise.}
\end{figure}

\newpage
\begin{figure}
\includegraphics[width=0.75\textwidth]{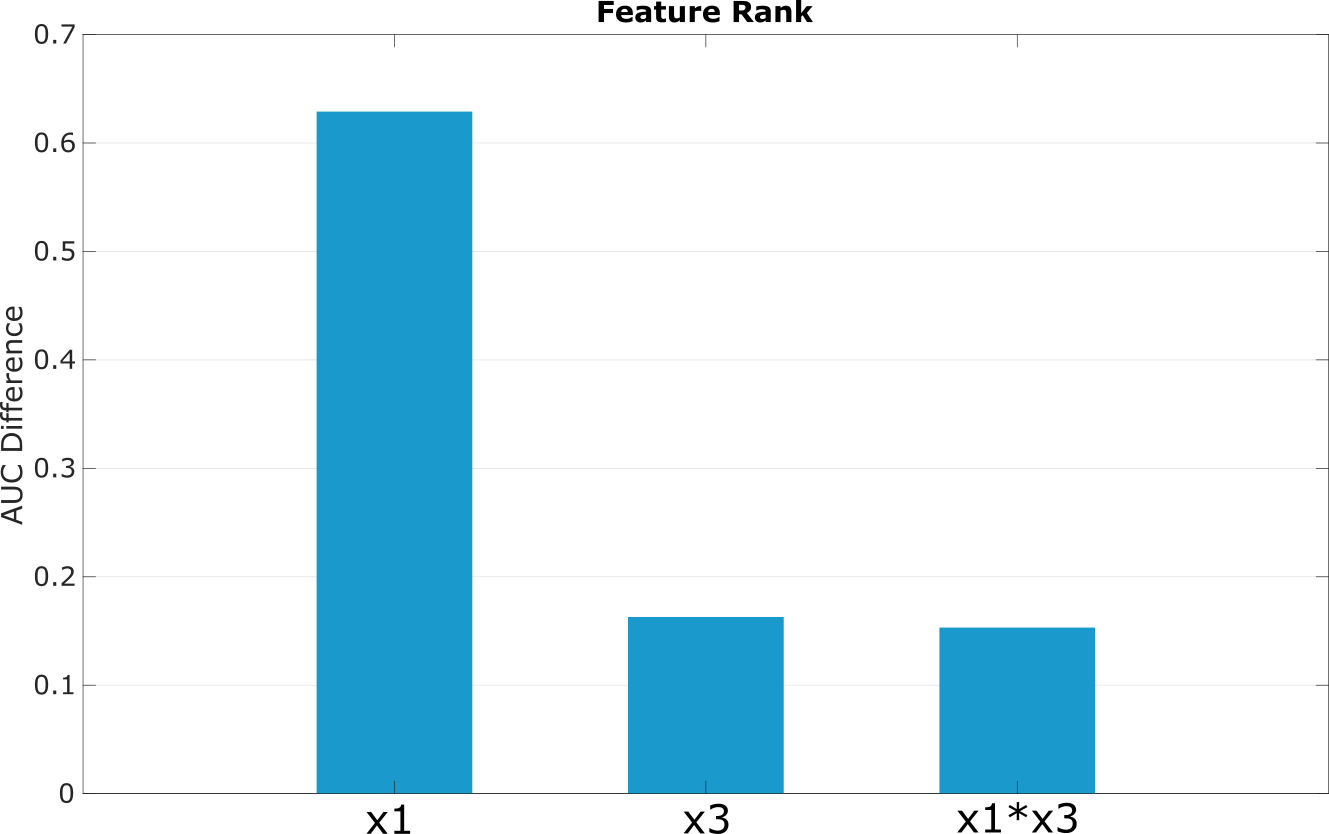}
\caption{\label{fig:k4} Toy application 2: Feature ranking.}
\end{figure}

\newpage
\begin{figure}
\includegraphics[width=0.75\textwidth]{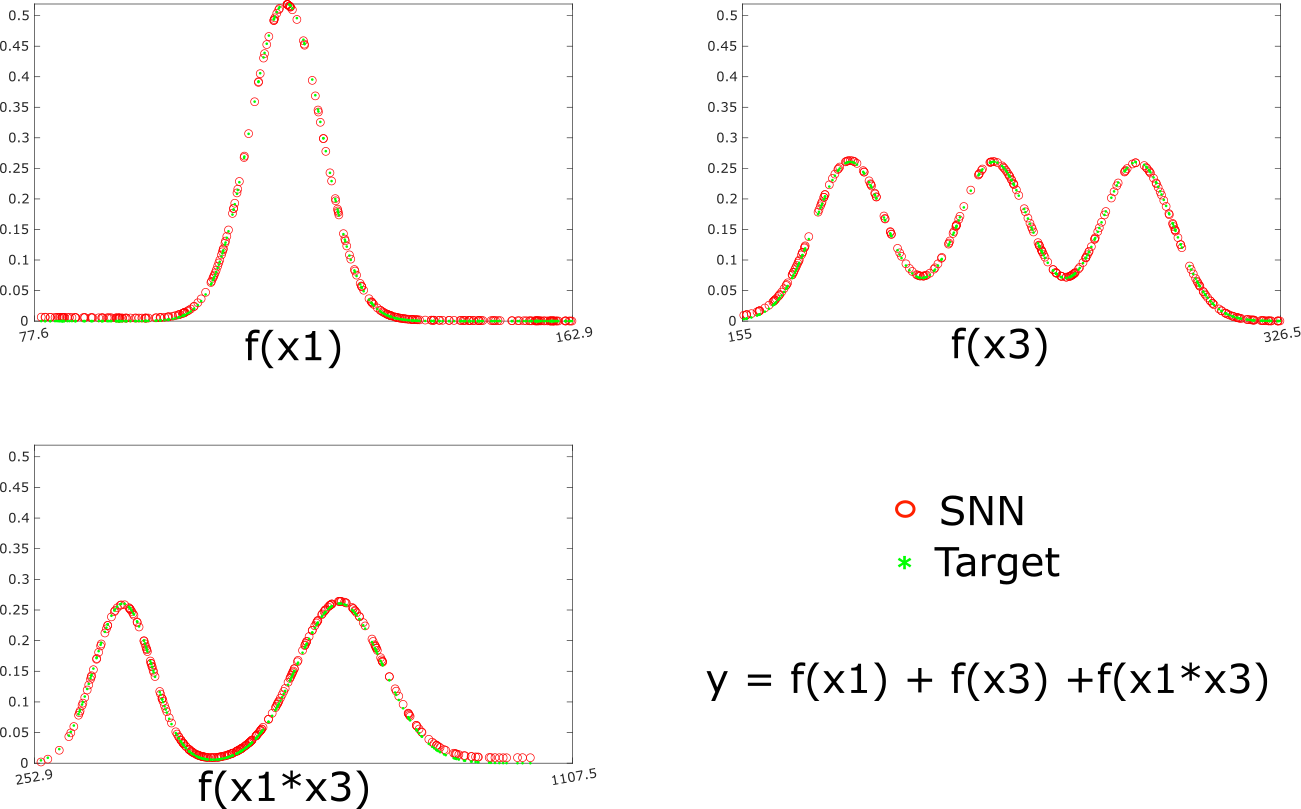}
\caption{\label{fig:k5} Toy application 2: Individual feature functions target output v.s. SNN output.}
\end{figure}

\FloatBarrier

\begin{table}
\renewcommand{\tablename}{{\bf Supplementary Table}}
\renewcommand{\thetable}{{\bf 1}}
\centering
\includegraphics[width=1.0\textwidth]{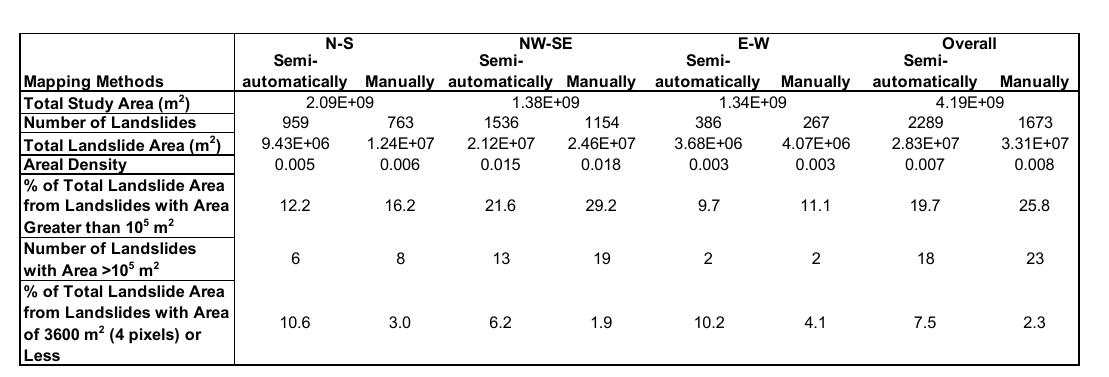}
\caption{\label{tab:S1} Description of Landslide Inventory.}
\end{table}

\begin{table}
  \renewcommand{\tablename}{{\bf Supplementary Table}}
  \renewcommand\thetable{{\bf 2}}
\centering
\includegraphics[width=1.1\textwidth]{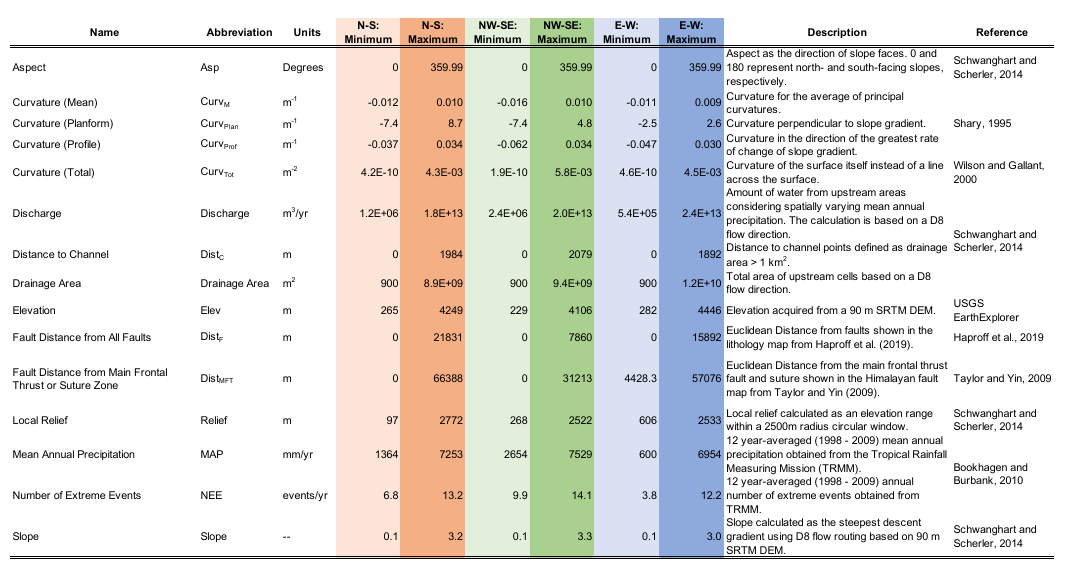}
\caption{\label{tab:S2} Description and Ranges of 15 Features.}
\end{table}

\begin{table}
  \renewcommand{\tablename}{{\bf Supplementary Table}}
  \renewcommand\thetable{{\bf 3}}
\centering
\includegraphics[width=0.9\textwidth]{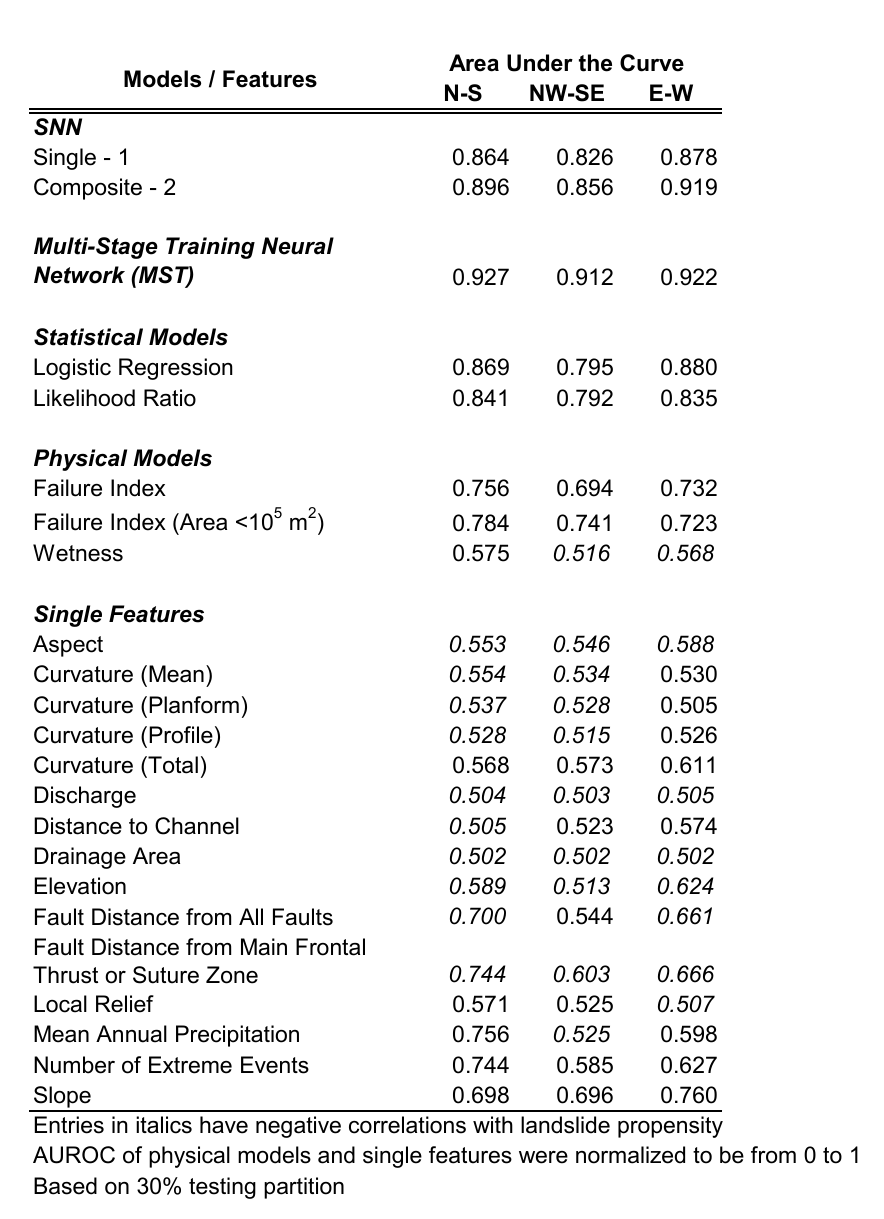}
\caption{\label{tab:S3} AUROC of Models and Single Features.}
\end{table}

\begin{table}
  \renewcommand{\tablename}{{\bf Supplementary Table}}
  \renewcommand\thetable{{\bf 4}}
\centering
\includegraphics[width=0.8\textwidth]{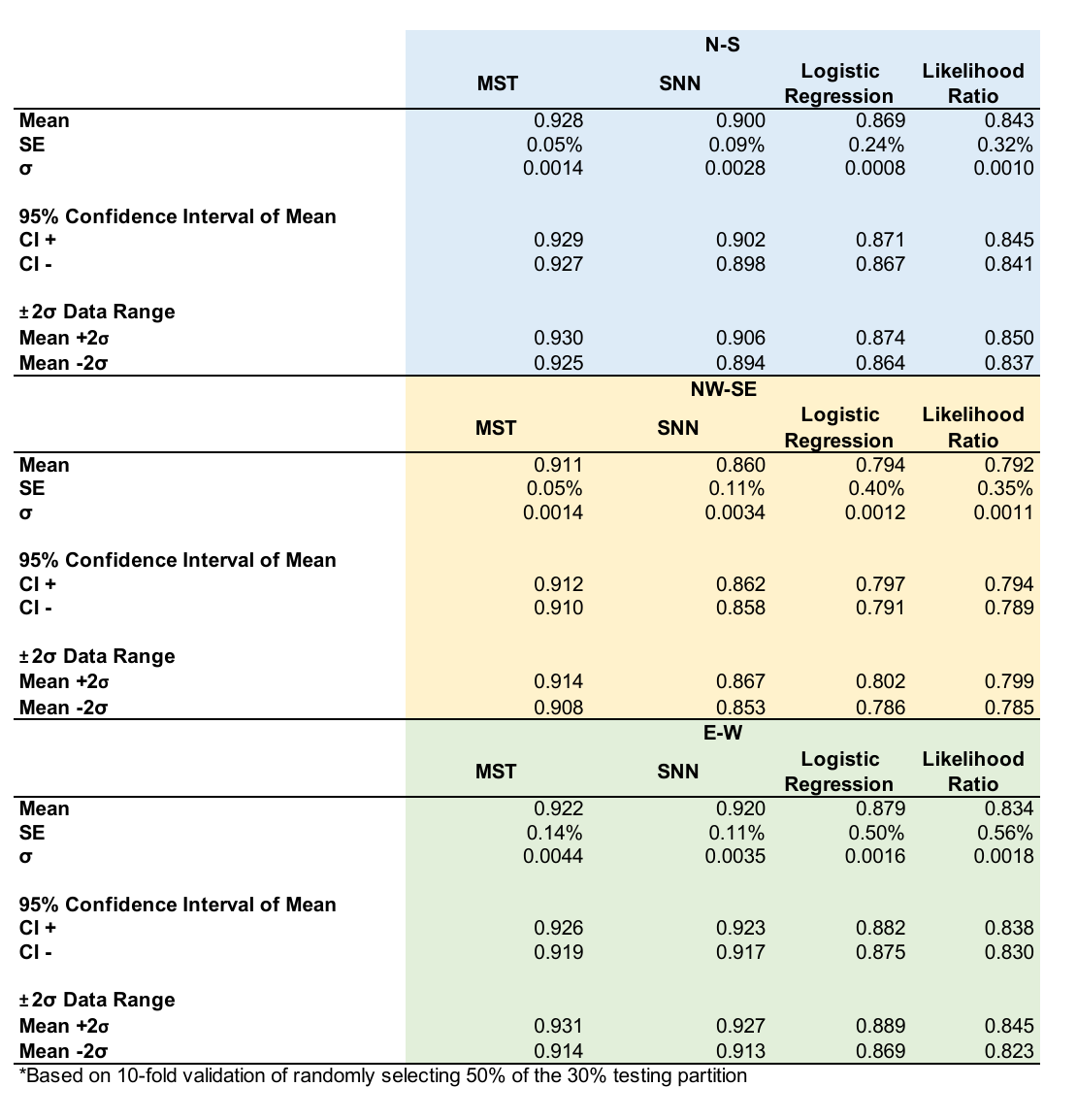}
\caption{\label{tab:S4} Artificial Neural Network and Statistical Model Confidence Intervals.}
\end{table}

\begin{table}
  \renewcommand{\tablename}{{\bf Supplementary Table}}
  \renewcommand\thetable{{\bf 5}}
\centering
\includegraphics[width=0.9\textwidth]{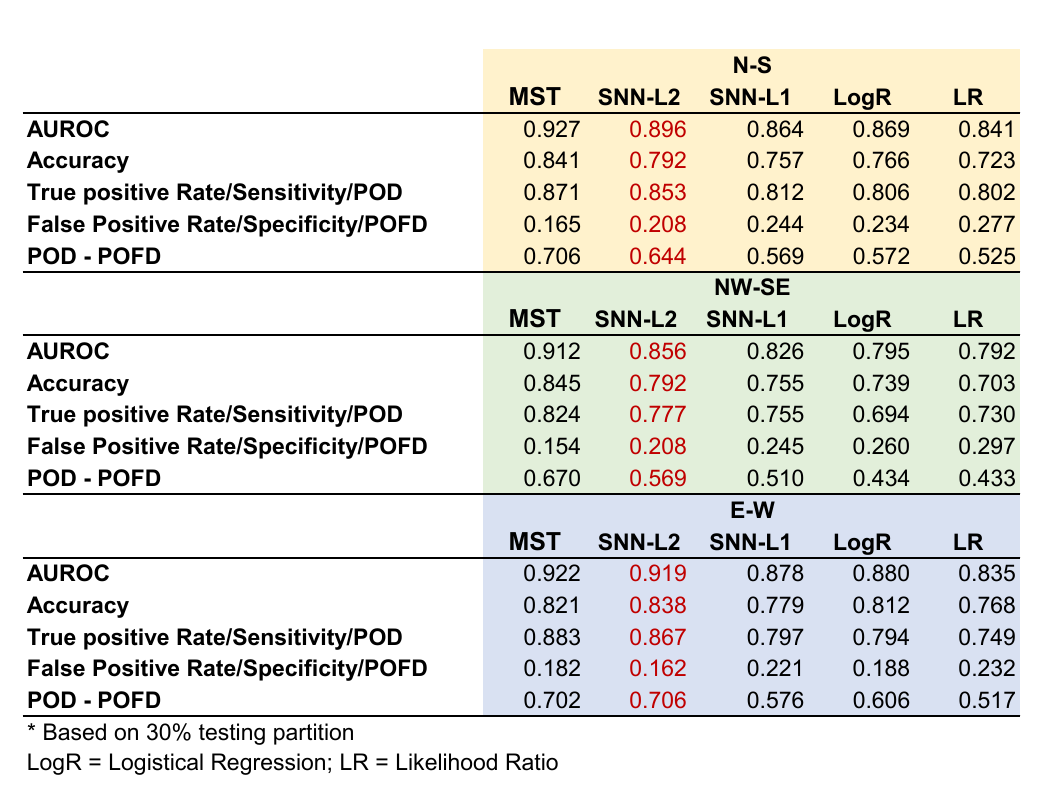}
\caption{\label{tab:S5} Performance Metrics for Models.}
\end{table}

\begin{table}
\renewcommand{\tablename}{{\bf Supplementary Table}}
\renewcommand{\thetable}{{\bf 6}}
\centering
\begin{tabular}{cccccccc}
$x_1$ & $x_2$ & $x_3$ & $x_4$ & $x_1*x_2$ & $x_3*x_4$ & $x_1*x_2*x_3*x_4$ & $y$ \\
0 & 0	& 0 & 0 & 0 & 0	& 0 & 0 \\
0 & 0	& 0 & 1 & 0 & 0	& 0 & 0 \\
0 & 0	& 1 & 0 & 0 & 0	& 0 & 0 \\
0 & 0	& 1 & 1 & 0 & 1	& 0 & 1 \\
0 & 1	& 0 & 0 & 0 & 0	& 0 & 0 \\
0 & 1	& 0 & 1 & 0 & 0	& 0 & 0 \\
0 & 1	& 1 & 0 & 0 & 0	& 0 & 0 \\
0 & 1	& 1 & 1 & 0 & 1	& 0 & 1 \\
1 & 0	& 0 & 0 & 0 & 0	& 0 & 0 \\
1 & 0	& 0 & 1 & 0 & 0	& 0 & 0 \\
1 & 0	& 1 & 0 & 0 & 0	& 0 & 0 \\
1 & 0	& 1 & 1 & 0 & 1	& 0 & 1 \\
1 & 1	& 0 & 0 & 1 & 0	& 0 & 1 \\
1 & 1	& 0 & 1 & 1 & 0	& 0 & 1 \\
1 & 1	& 1 & 0 & 1 & 0	& 0 & 1 \\
1 & 1	& 1 & 1 & 1 & 1	& 1 & 0
\end{tabular}
\caption{\label{tab:1} Truth table.}
\end{table}

\begin{table}
\renewcommand{\tablename}{ {\bf Supplementary Table} }
\renewcommand{\thetable}{{\bf 7}}
\centering
\begin{tabular}{cc}
Feature & Level \\
$x_1$ & 1 \\
$x_2$ & 1 \\
$x_3$ & 1 \\
$x_4$ & 1 \\
$x_1*x_2$ & 2 \\
$x_1*x_3$ & 2 \\
$x_1*x_4$ & 2 \\
$x_2*x_3$ & 2 \\
$x_2*x_4$ & 2 \\
$x_3*x_4$ & 2 \\
$x_1*x_2*x_3$ & 3 \\
$x_1*x_2*x_4$ & 3 \\
$x_1*x_3*x_4$ & 3 \\
$x_2*x_3*x_4$ & 3 \\
$x_1*x_2*x_3*x_4$ & 4
\end{tabular}
\caption{\label{tab:2} Composite features.}
\end{table}

\begin{table}
  \renewcommand{\tablename}{{\bf Supplementary Table}}
  \renewcommand\thetable{{\bf 8}}
\centering
\includegraphics[width=1.1\textwidth]{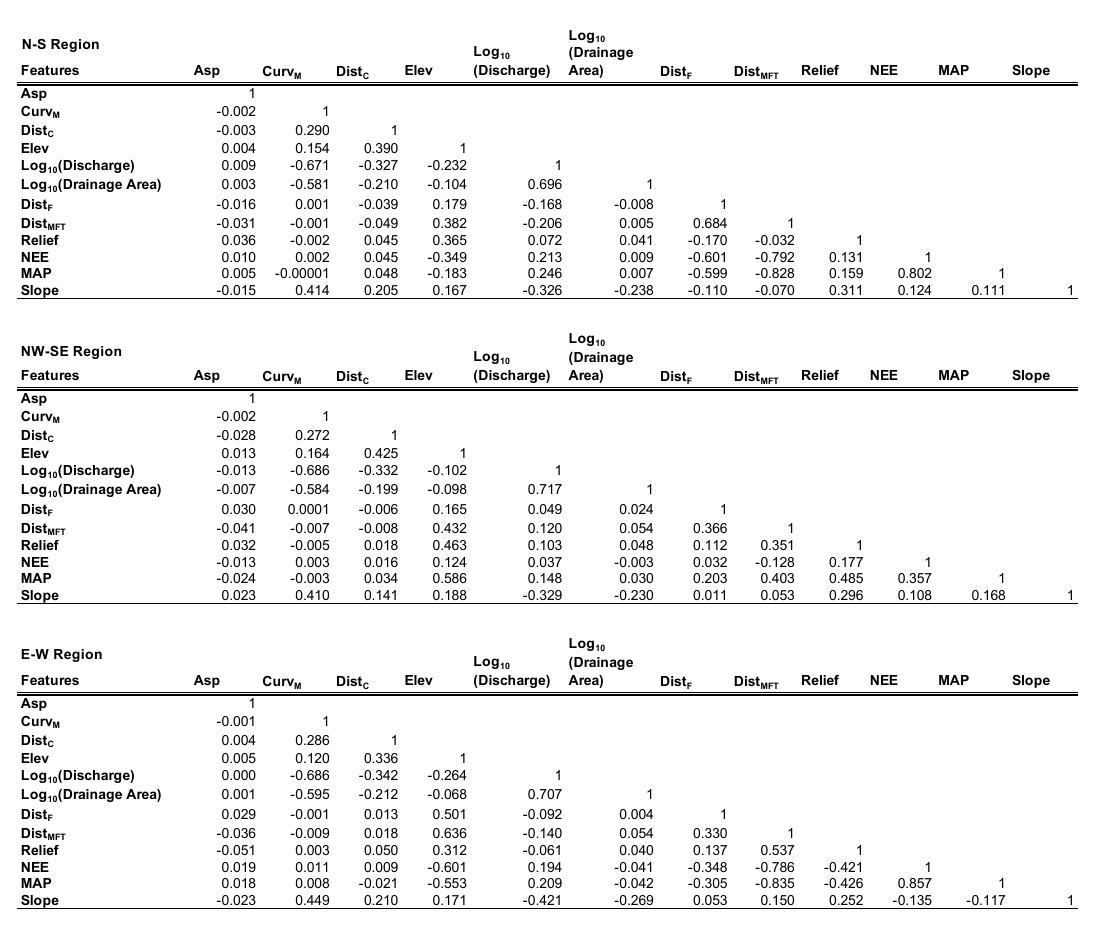}
\caption{\label{tab:S6} Correlation Metrics Between Features (R-value).}
\end{table}

\begin{table}
  \renewcommand{\tablename}{{\bf Supplementary Table}}
  \renewcommand\thetable{{\bf 9}}
\centering
\includegraphics[width=0.9\textwidth]{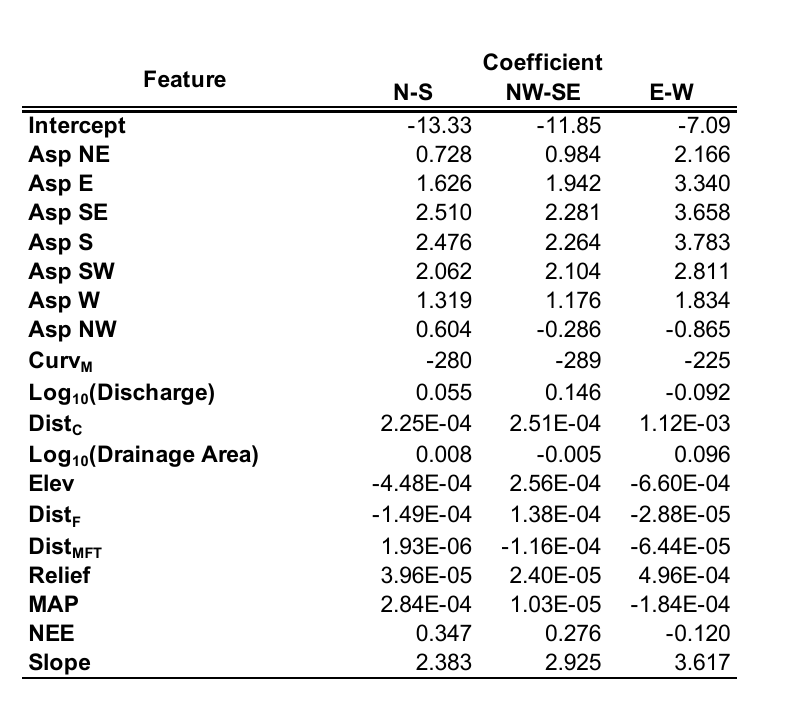}
\caption{\label{tab:S7} Logistic Regression Control Coefficients.}
\end{table}

\FloatBarrier

\newpage

\bibliographystyle{unsrt}

\bibliography{ldref_CEE_KS.bib}

\end{document}